\documentclass[10pt,doublespacing]{elsart}
\usepackage{latexsym,amssymb,amsmath,graphicx,epsfig,cite,bbm,float,epsf,graphics}

\journal{Digital Signal Processing}

\newcommand{\nn}{\nonumber}

\renewcommand{\vec}[1]{\mbox{\boldmath${#1}$}}

\newcommand{\ei}{\end{itemize}}
\newcommand{\bi}{\begin{itemize}}

\newcommand{\vu}{\mbox{$\vec{u}$}}
\newcommand{\vl}{\vec{\lambda}}

\newcommand{\vgamma}{\vec{\gamma}}
\newcommand{\vGamma}{\vec{\Gamma}}

\newcommand{\tr}{\mathrm{tr}}

\newcommand{\limitt}{\lim_{t\rightarrow \infty}}

\newcommand{\vw}{\vec{w}}
\newcommand{\vW}{\mbox{$\vec{W}$}}

\newcommand{\va}{\mbox{$\vec{a}$}}

\newcommand{\vx}{\mbox{$\vec{x}$}}
\newcommand{\vR}{\mbox{$\vec{R}$}}
\newcommand{\vp}{\mbox{$\vec{p}$}}
\newcommand{\vz}{\mbox{$\vec{z}$}}
\newcommand{\vv}{\mbox{$\vec{v}$}}
\newcommand{\vq}{\mbox{$\vec{q}$}}
\newcommand{\vQ}{\mbox{$\vec{Q}$}}

\newcommand{\vf	}{\vec{f}}

\newcommand{\MB}{\left[\begin{array}}
\newcommand{\ME}{\end{array}\right]}

\newcommand{\veps}{\mbox{$\vec{\varepsilon}$}}

\newcommand{\defi}{\stackrel{\bigtriangleup}{=}}
\makeatletter

\newcommand{\Rmnum}[1]{\expandafter\@slowromancap\romannumeral #1@}
\makeatother

\def\ninept{\def\baselinestretch{1.5}}
\ninept

\begin{document}
\begin{frontmatter}

\title{Adaptive Mixture Methods Based on Bregman Divergences}
\vspace{-0.5in}
\author[1]{Mehmet A. Donmez},
\ead{mdonmez@ku.edu.tr}
\author[1]{Huseyin A. Inan},
\ead{huseyin.inan@boun.edu.tr}
\author[1]{Suleyman S. Kozat\corauthref{cor}}
\corauth[cor]{Corresponding author.}
\ead{skozat@ku.edu.tr}
\address[1]{Department of Electrical and Computer Engineering, Koc University, Istanbul, Tel: 90 212 3501840.}
\vspace{-0.3in}
\begin{abstract}
We investigate adaptive mixture methods that linearly combine outputs
of $m$ constituent filters running in parallel to model a desired
signal. We use ``Bregman divergences'' and obtain certain
multiplicative updates to train the linear combination weights under
an affine constraint or without any constraints. We use unnormalized
relative entropy and relative entropy to define two different Bregman
divergences that produce an unnormalized exponentiated gradient
update and a normalized exponentiated gradient update on the mixture
weights, respectively. We then carry out the mean and the mean-square
transient analysis of these adaptive algorithms when they are used to
combine outputs of $m$ constituent filters. We illustrate the accuracy
of our results and demonstrate the effectiveness of these updates for
sparse mixture systems.
\end{abstract}
\begin{keyword}
Adaptive mixture, Bregman divergence, affine mixture, multiplicative update.
\end{keyword}
\end{frontmatter}
\section{Introduction} 
In this paper, we study adaptive mixture methods based on ``Bregman
divergences'' \cite{dsp_3, multip1} that combine outputs of $m$ constituent
filters running in parallel on the same task. The overall system has
two stages \cite{dsp_1, dsp_2, kozat2, convex3, convex4, GaGoVi05}. The
first stage contains adaptive filters running in parallel to model a
desired signal. The outputs of these adaptive filters are then
linearly combined to produce the final output of the overall system in
the second stage.  We use Bregman divergences and obtain certain
multiplicative updates \cite{EG}, \cite{multip1}, \cite{multip2} to
train these linear combination weights under an affine constraint
\cite{bershard} or without any constraints \cite{kozat}. We use
unnormalized \cite{multip1} and normalized relative entropy \cite{EG}
to define two different Bregman divergences that produce the
unnormalized exponentiated gradient update (EGU) and the exponentiated
gradient update (EG) on the mixture weights \cite{EG}, respectively. We then
perform the mean and the mean-square transient analysis of these
adaptive mixtures when they are used to combine outputs of $m$
constituent filters. We emphasize that to the best of our knowledge,
this is the first mean and mean-square transient analysis of the EGU
algorithm and the EG algorithm in the mixture framework (which
naturally covers the classical framework also
\cite{EGvsLMS,sayed}). We illustrate the accuracy of our results
through simulations in different configurations and demonstrate
advantages of the introduced algorithms for sparse mixture systems.

Adaptive mixture methods are utilized in a wide range of signal
processing applications in order to improve the steady-state and/or
convergence performance over the constituent filters
\cite{bershard,kozat,convex}. An adaptive convexly constrained mixture
of two filters is studied in \cite{convex}, where the convex
combination is shown to be ``universal'' such that the combination
performs at least as well as its best constituent filter in the
steady-state \cite{convex}. The transient analysis of this adaptive
convex combination is studied in \cite{transientconvex}, where the
time evolution of the mean and variance of the mixture weights is
provided. In similar lines, an affinely constrained mixture of
adaptive filters using a stochastic gradient update is introduced in
\cite{bershard}. The steady-state mean square error (MSE) of this
affinely constrained mixture is shown to outperform the steady-state
MSE of the best constituent filter in the mixture under certain
conditions \cite{bershard}. The transient analysis of this affinely
constrained mixture for $m$ constituent filters is carried out in
\cite{transientaffine}. The general linear mixture framework as well
as the steady-state performances of different mixture configurations
are studied in \cite{kozat}.

In this paper, we use Bregman divergences to derive multiplicative
updates on the mixture weights. We use the unnormalized relative
entropy and the relative entropy as distance measures and obtain the
EGU algorithm and the EG algorithm to update the combination weights
under an affine constraint or without any constraints. We then carry
out the mean and the mean-square transient analysis of these adaptive
mixtures when they are used to combine $m$ constituent filters. We
point out that the EG algorithm is widely used in sequential learning
theory \cite{vovk} and minimizes an approximate final estimation error
while penalizing the distance between the new and the old filter
weights. In network and acoustic echo cancellation applications, the
EG algorithm is shown to converge faster than the LMS algorithm
\cite{sayed,dsp_4} when the system impulse response is sparse
\cite{EGvsLMS}. Similarly, in our simulations, we observe that using
the EG algorithm to train the mixture weights yields increased
convergence speed compared to using the LMS algorithm to train the
mixture weights \cite{kozat,bershard} when the combination favors only
a few of the constituent filters in the steady state, i.e., when the
final steady-state combination vector is sparse.  We also observe that
the EGU algorithm and the LMS algorithm show similar performance when
they are used to train the mixture weights even if the final
steady-state mixture is sparse.

To summarize, the main contributions of this paper are
as follows:
\begin{itemize}
\item We use Bregman divergences to derive multiplicative updates on
  affinely constrained and unconstrained mixture weights adaptively
  combining outputs of $m$ constituent filters.
\item We use the unnormalized relative entropy and the relative entropy to define two different Bregman divergences that produce the EGU algorithm and the EG algorithm to update the affinely constrained and unconstrained mixture weights.
\item We perform the mean and the mean-square transient analysis of the affinely constrained and unconstrained mixtures using the EGU algorithm and the EG algorithm.
\end{itemize}

The organization of the paper is as follows. In Section \Rmnum{2}, we
first describe the mixture framework. In Section \Rmnum{3}, we study
the affinely constrained and unconstrained mixture methods updated
with the EGU algorithm and the EG algorithm. In Section \Rmnum{4}, we
first perform the transient analysis of the affinely constrained
mixtures and then continue with the transient analysis of the
unconstrained mixtures. Finally, in Section \Rmnum{5}, we perform
simulations to show the accuracy of our results and to compare
performances of the different adaptive mixture methods. The paper
concludes with certain remarks in Section \Rmnum{6}.
\section{System Description} 
\subsection{Notation}
In this paper, all vectors are column vectors and represented by
boldface lowercase letters. Matrices are represented by boldface
capital letters. For presentation purposes, we work only with real
data. Given a vector $\vw$, $w^{(i)}$ denotes the $i$th individual
entry of $\vw$, $\vw^T$ is the transpose of $\vw$, $\|\vw\|_1 \defi
\sum_i |w^{(i)}|$ is the $l_1$ norm; $\|\vw\| \defi \sqrt{\vw^T\vw}$
is the $l_2$ norm. For a matrix $\vW$, $\tr(\vW)$ is the trace.  For a vector $\vw$, $\mathrm{diag}(\vw)$
represents a diagonal matrix formed using the entries of $\vw$. For a matrix $\vW$, $\mathrm{diag}(\vW)$ represents a column vector that contains the diagonal entries of $\vW$. For
two vectors $\vv_1$ and $\vv_2$, we define the concatenation
$[\vv_1;\vv_2] \defi [\vv_1^T\;\vv_2^T]^T$.  For a random variable
$v$, $\bar{v}$ is the expected value. For a random vector $\vv$ (or a
random matrix $\vec{V}$), $\bar{\vv}$ (or $\bar{\vec{V}}$) represents
the expected value of each entry. Vectors (or matrices) $\vec{1}$ and $\vec{0}$, with an abuse of
notation, denote vectors (or matrices) of all ones or zeros,
respectively, where the size of the vector (or the matrix) is
understood from the context.
\subsection{System Description}
\begin{figure}[t]
\centerline{\epsfxsize=13cm \epsfbox{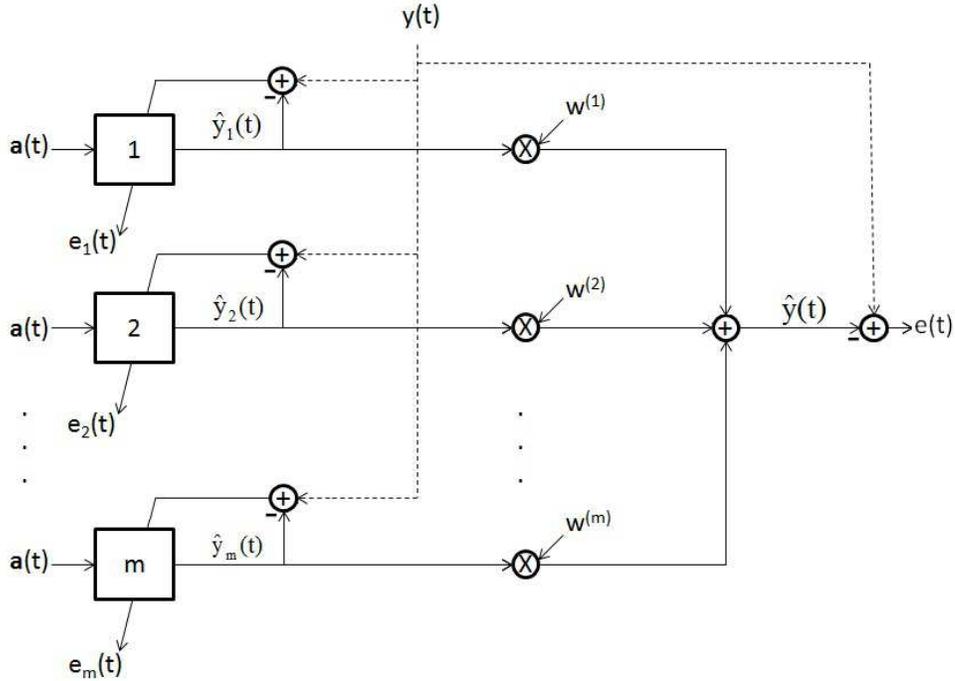} }
\caption{A linear mixture of outputs of $m$ adaptive filters.}
\label{fig:11}
\end{figure}
The framework that we study has two stages. In the first stage, we
have $m$ adaptive filters producing outputs $\hat{y}_i(t)$,
$i=1,\ldots,m$, running in parallel to model a desired signal $y(t)$
as seen in Fig. 1. The second stage is the mixture stage, where the
outputs of the first stage filters are combined to improve the
steady-state and/or the transient performance over the constituent
filters. We linearly combine the outputs of the first stage filters to
produce the final output as $\hat{y}(t) = \vw^T(t)\vx(t)$, where
$\vx(t) \defi [\hat{y}_1(t),\ldots,\hat{y}_m(t)]^T$ and train the
mixture weights using multiplicative updates (or exponentiated
gradient updates) \cite{multip1}. We point out that in order to
satisfy the constraints and derive the multiplicative updates
\cite{EG}, \cite{cesab}, we use reparametrization of the mixture
weights as $\vw(t) = \vf(\vz(t))$ and perform the update on $\vz(t)$
as
\begin{equation}
 \vz(t+1) = \arg \min_{\vz} \bigg\{
d(\vz,\vz(t)) + \mu \; l\big(y(t),\vf^T(\vz) \vx(t)\big)
\bigg\}, \label{eq:mul}
\end{equation}
where $\mu$ is the learning rate of the update, $d(\cdot,\cdot)$ is an appropriate distance measure and $l(\cdot,\cdot)$ is the instantaneous loss. We emphasize that in \eqref{eq:mul}, the updated vector $\vz$ is forced
to be close to the present vector $\vz(t)$ by $d(\vz(t+1),\vz(t))$, while trying to accurately model the current data by $ l\big(y(t),\vf^T(\vz) \vx(t)\big)$. However, instead of directly minimizing \eqref{eq:mul}, a linearized version of \eqref{eq:mul}
\begin{align}
& \vz(t+1) = \arg \min_{\vz} \bigg\{ d(\vz,\vz(t)) + l\left(y(t),\vf^T(\vz(t)) \vx(t)\right) \nn \\ & +
\mu \nabla_{\vz} l\left(y(t),\vf^T(\vz) \vx(t)\right)^T
\Big|_{\vz=\vz(t)} (\vz-\vz(t)) \bigg\} \label{eq:mul2}
\end{align}
is minimized to get the desired update. As an example, if we use the $l_2$-norm as the distance measure, i.e., $d(\vz,\vz(t)) = \|\vz-\vz(t)\|^2$, and the square error as the instantaneous loss, i.e.,  $l\big(y(t),\vf^T(\vz) \vx(t)\big)=[y(t)-\vf^T(\vz) \vx(t)]^2$ with
$\vf(\vz) = \vz$, then we get the stochastic gradient update on $\vw(t)$,
i.e.,
\[
\vw(t+1) = \vw(t) + \mu e(t) \vx(t),
\]
in \eqref{eq:mul2}.

In the next section, we use the unnormalized relative entropy
\begin{equation}
 d_1(\vz,\vz(t)) = \bigg\{ \sum_{i=1}^m \left[ z^{(i)} \ln
   \left(\frac{z^{(i)}}{z^{(i)}(t)} \right) + z^{(i)}(t) - z^{(i)}
   \right] \bigg\} \label{eq:mul_dist}
\end{equation}
for positively constrained $\vz$ and $\vz(t)$,
$\vz \in \mathbbm{R}_+^m$, $\vz(t) \in \mathbbm{R}_+^m$, and the relative
entropy
\begin{equation}
 d_2(\vz,\vz(t)) = \bigg\{ \sum_{i=1}^m \left[ z^{(i)} \ln
   \left(\frac{z^{(i)}}{z^{(i)}(t)} \right)
   \right] \bigg\}, \label{eq:mul_dist2}
\end{equation}
where $\vz$ is constrained to be in an extended simplex such that
$z^{(i)} \geq 0$, $\sum_{k=1}^m z^{(i)} = u$ for some $u\geq1$ as the
distance measures, with appropriately selected $\vf(\cdot)$ to derive
updates on mixture weights under different constraints. We first investigate affinely constrained mixture of $m$ adaptive filters, and then continue with the unconstrained mixture using \eqref{eq:mul_dist} and \eqref{eq:mul_dist2} as the distance measures.

\section{Adaptive Mixture Algorithms} 
In this section, we investigate affinely constrained and unconstrained mixtures updated with the EGU algorithm and the EG algorithm.

\subsection{Affinely Constrained Mixture}
When an affine constraint is imposed on the mixture such that $\vw^T(t) \vec{1}=1$, we get
\begin{align*}
& \hat{y}(t)  = \vw(t)^T\vx(t), \\
& e(t)  = y(t) -\hat{y}(t), \\
& w^{(i)}(t) = \lambda^{(i)}(t), \: \: i=1,\ldots,m-1, \\
& w^{(m)}(t) = 1- \sum_{i=1}^{m-1} \lambda^{(i)}(t),
\end{align*}
where the $m-1$ dimensional vector $\vl(t)\defi
[\lambda^{(1)}(t),\ldots,\lambda^{(m-1)}(t)]^T$ is the unconstrained
weight vector, i.e., $\vl(t) \in \mathbbm{R}^{m-1}$. Using $\vl(t)$ as
the unconstrained weight vector, the error can be written as $e(t) =
\big[ y(t)-\hat{y}_m(t) \big] - \vl^T(t) \vec{\delta}(t)$, where
$\vec{\delta}(t) \defi [\hat{y}_1(t)-\hat{y}_m(t),\ldots,\hat{y}_{m-1}(t)-\hat{y}_m(t)]^T$.
To be able to derive a multiplicative update on $\vl(t)$, we use
\[
\vl(t)  = \vl_1(t)-\vl_2(t),
\]
where $\vl_1(t)$ and $\vl_2(t)$ are constrained to be nonnegative,
i.e., $\vl_i(t)\in \mathbbm{R}_+^{m-1}$, $i=1,2$. After we collect unconstrained weights in $\vl_a(t)=[\vl_1(t);\vl_2(t)]$, we define a function of loss $e(t)$ as
\[
l_{\mbox{a}}\left(\vl_a(t)\right) \defi e^2(t)
\]
and update positively constrained $\vl_a(t)$ as follows.
\subsubsection{Unnormalized Relative Entropy}
Using the unconstrained relative entropy as the distance measure, we get
\begin{align}
\vl_a(t+1) = & \arg \min_{\vl} \bigg\{ \sum_{i=1}^{2(m-1)} \left[ \lambda^{(i)} \ln \left( \frac{\lambda^{(i)}}{\lambda_a^{(i)}(t)} \right) + \lambda_a^{(i)}(t)-\lambda^{(i)} \right] + \nn \\
& \mu  \left[ l_{\mbox{a}}\left(\vl_a(t)\right)+ \nabla_{\vl} l_{\mbox{a}}\left(\vl\right)^T \big|_{\vl = \vl_a(t)} \left(\vl - \vl_a(t)\right) \right] \bigg\}. \nn
\end{align}
After some algebra this yields
\begin{align}
& \lambda_a^{(i)}(t+1) = \lambda_a^{(i)}(t) \exp\left\{  \mu e(t)  (\hat{y}_i(t)-\hat{y}_m(t)) \right\}, i = 1,\ldots , m-1, \nn \\
& \lambda_a^{(i)}(t+1) = \lambda_a^{(i)}(t) \exp\left\{ -  \mu e(t)  (\hat{y}_i(t)-\hat{y}_m(t)) \right\}, i = m, \ldots, 2(m-1), \nn
\end{align}
providing the multiplicative updates on $\vl_1(t)$ and $\vl_2(t)$.

\subsubsection{Relative Entropy}
Using the relative entropy as the distance measure, we get
\begin{align}
\vl_a(t+1) = & \arg \min_{\vl} \bigg\{ \sum_{i=1}^{2(m-1)} \left[ \lambda^{(i)} \ln \left( \frac{\lambda^{(i)}}{\lambda_a^{(i)}(t)} \right) + \gamma(u-\vec{1}^T \vl) \right] + \nn \\
& \mu  \left[ l_{\mbox{a}}\left(\vl_a(t)\right)+ \nabla_{\vl} l_{\mbox{a}}\left(\vl\right)^T \big|_{\vl = \vl_a(t)} \left(\vl - \vl_a(t)\right) \right] \bigg\}, \nn
\end{align}
where $\gamma$ is the Lagrange multiplier. This yields \scriptsize
\begin{align*}
& \lambda_a^{(i)}(t+1) = u \frac{\lambda_a^{(i)}(t) \exp\left\{  \mu e(t)  (\hat{y}_i(t)-\hat{y}_m(t)) \right\}}{\sum_{k=1}^{m-1}\bigg[ \lambda_a^{(k)}(t) \exp\left\{  \mu e(t)  (\hat{y}_k(t)-\hat{y}_m(t))\right\} + \lambda_a^{(k+m-1)}(t) \exp\left\{ -  \mu e(t)  (\hat{y}_k(t)-\hat{y}_m(t))\right\}\bigg] }, \nn \\
& i=1,\ldots,m-1, \nn \\
& \lambda_a^{(i)}(t+1) = u \frac{\lambda_a^{(i)}(t) \exp\left\{ -  \mu e(t)  (\hat{y}_i(t)-\hat{y}_m(t)) \right\}}{\sum_{k=1}^{m-1}\bigg[ \lambda_a^{(k)}(t) \exp\left\{  \mu e(t)  (\hat{y}_k(t)-\hat{y}_m(t))\right\} + \lambda_a^{(k+m-1)}(t) \exp\left\{ -  \mu e(t)  (\hat{y}_k(t)-\hat{y}_m(t))\right\}\bigg]}, \nn \\
& i=m,\ldots,2(m-1), \nn
\end{align*} \normalsize
providing the multiplicative updates on $\vl_a(t)$.

\subsection{Unconstrained Mixture}
Without any constraints on the combination weights, the mixture stage can be written as
\begin{align*}
& \hat{y}(t) = \vw^T(t)\vx(t), \\
& e(t) = y(t) -\hat{y}(t),
\end{align*}
where $\vw(t) \in \mathbbm{R}^{m}$. To be able to derive a multiplicative update, we use a change of variables,
\[
\vw(t) = \vw_1(t)-\vw_2(t),
\]
where $\vw_1(t)$ and $\vw_2(t)$ are constrained to be nonnegative,
i.e., $\vw_i(t) \in \mathbbm{R}^m_+$, $i=1,2$.  We then collect the unconstrained weights $\vw_a(t) = [\vw_1(t);\vw_2(t)]$ and define a function of the loss $e(t)$ as
\[
l_{\mbox{u}}\left(\vw_a(t)\right) \defi e^2(t).
\]
\subsubsection{Unnormalized Relative Entropy}
Defining cost function similar to (4) and minimizing it with respect to $\vw$ yields
\begin{align}
& w_a^{(i)}(t+1) = w_a^{(i)}(t) \exp\left\{  \mu e(t)  \hat{y}_i(t) \right\}, \nn i= 1, \ldots , m,\\
& w_a^{(i)}(t+1) = w_a^{(i)}(t) \exp\left\{ -  \mu e(t)  \hat{y}_i(t) \right\}, \nn i = m+1, \ldots , 2m,
\end{align}
providing the multiplicative update on $\vw_a(t)$.

\subsubsection{Relative Entropy}
Using the relative entropy under the simplex constraint on $\vw$, we get the updates
\begin{align}
& w_a^{(i)}(t+1) = u \frac{w_a^{(i)}(t) \exp\left\{  \mu e(t)  \hat{y}_i(t) \right\}}{\displaystyle \sum_{k=1}^m \bigg[ w_a^{(k)}(t) \exp\left\{  \mu e(t)  \hat{y}_k(t) \right\} + w_a^{(k+m)}(t) \exp\left\{ - \mu e(t)  \hat{y}_k(t) \right\} \bigg]}, \nn \\
& i=1,\ldots,m, \nn \\
& w_a^{(i)}(t+1) = u \frac{w_a^{(i)}(t) \exp\left\{ -  \mu e(t)  \hat{y}_i(t) \right\}}{\displaystyle \sum_{k=1}^m \bigg[ w_a^{(k)}(t) \exp\left\{  \mu e(t)  \hat{y}_k(t) \right\} + w_a^{(k+m)}(t) \exp\left\{ - \mu e(t)  \hat{y}_k(t) \right\} \bigg]}, \nn \\
& i=m+1\ldots,2m. \nn
\end{align}

In the next section, we study the transient analysis of these four
adaptive mixture algorithms.
 
\section{Transient Analysis} 
In this section, we study the mean and the mean-square transient
analysis of the adaptive mixture methods. We start with the affinely
constrained combination.
\subsection{Affinely Constrained Mixture}
We first perform the transient analysis of the mixture weights updated
with the EGU algorithm. Then, we continue with the transient analysis
of the mixture weights updated with the EG algorithm.
\subsubsection{Unconstrained Relative Entropy}
For the affinely constrained mixture updated with the EGU algorithm, we have the multiplicative update as
\begin{align}
 \lambda_1^{(i)}(t+1) & = \lambda_1^{(i)}(t) \exp\left\{  \mu e(t)  (\hat{y}_i(t)-\hat{y}_m(t)) \right\}, \nn \\
               & = \lambda_1^{(i)}(t) \sum_{k=0}^{\infty} \frac{ \big( \mu e(t) (\hat{y}_i(t)-\hat{y}_m(t)) \big)^k}{k!}, \label{eq:lambda1} \\
 \lambda_2^{(i)}(t+1) & = \lambda_2^{(i)}(t) \exp\left\{ -  \mu e(t)  (\hat{y}_i(t)-\hat{y}_m(t)) \right\}, \nn \\
               & = \lambda_2^{(i)}(t) \sum_{k=0}^{\infty} \frac{ \big(- \mu e(t) (\hat{y}_i(t)-\hat{y}_m(t)) \big)^k}{k!}, \label{eq:lambda2}
\end{align}
for $i=1,\ldots,m-1$. If $e(t)$ and $\hat{y}_i(t)-\hat{y}_m(t)$ for each $i=1,\ldots,m-1$ are bounded, then we can write \eqref{eq:lambda1} and \eqref{eq:lambda2} as
\begin{align}
& \lambda_1^{(i)}(t+1) = \lambda_1^{(i)}(t) \big(1 +  \mu e(t)  (\hat{y}_i(t)-\hat{y}_m(t)) + O(\mu^2)\big) , \label{eq:un1} \\
& \lambda_2^{(i)}(t+1) = \lambda_2^{(i)}(t) \big(1 - \mu e(t) (\hat{y}_i(t)-\hat{y}_m(t)) + O(\mu^2)\big), \label{eq:un2}
\end{align}
for $i=1,\ldots,m-1$. Since $\mu$ is usually relatively small \cite{multip1}, we approximate \eqref{eq:un1} and \eqref{eq:un2} as
\begin{align}
& \lambda_1^{(i)}(t+1) = \lambda_1^{(i)}(t) \big(1 +  \mu e(t)  (\hat{y}_i(t)-\hat{y}_m(t)) \big) , \label{eq:un3} \\
& \lambda_2^{(i)}(t+1) = \lambda_2^{(i)}(t) \big(1 - \mu e(t) (\hat{y}_i(t)-\hat{y}_m(t)) \big). \label{eq:un4}
\end{align}
In our simulations, we illustrate the accuracy of the approximations \eqref{eq:un3} and \eqref{eq:un4} under the mixture framework.
Using \eqref{eq:un3} and \eqref{eq:un4}, we can obtain updates on $\vl_1(t)$ and $\vl_2(t)$ as
\begin{align}
& \vl_1(t+1) =  \big(I +  \mu e(t) \mathrm{diag}\big(\vec{\delta}(t)\big) \big)\vl_1(t) , \label{eq:un51} \\
& \vl_2(t+1) =  \big(I -  \mu e(t) \mathrm{diag}\big(\vec{\delta}(t)\big) \big)\vl_2(t). \label{eq:un61}
\end{align}
Collecting the weights in $\vl_a(t) = [\vl_1(t);\vl_2(t)]$, using the updates \eqref{eq:un51} and \eqref{eq:un61}, we can write update on $\vl_a(t)$ as
\begin{align}
\vl_a(t+1)= \big(I + \mu e(t) \mathrm{diag}\big(\vu(t)\big) \big)\vl_a(t), \label{eq:EGU1}
\end{align}
where $\vu(t)$ is defined as $\vu(t)\defi [\vec{\delta}(t);-\vec{\delta}(t)]$.

For the desired signal $y(t)$, we can write $ y(t)-\hat{y}_m(t) =
\vl_0^T(t) \vec{\delta}(t) + e_0(t)$, where $\vl_0(t)$ is the optimum
MSE solution at time $t$ such that $\vl_0(t) \defi \vR^{-1}(t)\vp(t)$,
$\vR(t) \defi E\big[\vec{\delta}(t)\vec{\delta}^T(t)\big]$, $\vp(t)
\defi E\left\{\vec{\delta}(t)\big[y(t)-\hat{y}_m(t)\big]\right\}$ and
$e_0(t)$ is zero-mean and uncorrelated with $\vec{\delta}(t)$. We next
show that the mixture weights converge to the optimum solution in the
steady-state such that $\limitt E\big[\vl(t)\big] = \limitt \vl_0(t)$
for properly selected $\mu$.

Subtracting \eqref{eq:un61} from \eqref{eq:un51}, we obtain
\begin{align}
\vl(t+1) & = \vl(t) + \mu  e(t) \mathrm{diag}\big(\vec{\delta}(t)\big) \big( \vl_1(t) + \vl_2(t) \big), \nn \\ & = \vl(t) - \mu  e(t) \mathrm{diag}\big(\vec{\delta}(t)\big) \vl(t) + 2 \mu  e(t) \mathrm{diag}\big(\vec{\delta}(t)\big) \vl_1(t). \label{eq:un7}
\end{align}
Defining $\veps(t) \defi \vl_0(t)- \vl(t)$ and using $e(t) = \vec{\delta}^T(t)\veps(t) + e_0(t)$ in \eqref{eq:un7} yield
\begin{align}
& \vl(t+1) = \vl(t) - \mu \mathrm{diag}\big(\vec{\delta}(t)\big) \vl(t) \vec{\delta}^T(t)\veps(t) - \mu \mathrm{diag}\big(\vec{\delta}(t)\big) \vl(t) e_0(t) \nn \\ & + 2 \mu \mathrm{diag}\big(\vec{\delta}(t)\big) \vl_1(t) \vec{\delta}^T(t)\veps(t) + 2 \mu \mathrm{diag}\big(\vec{\delta}(t)\big) \vl_1(t) e_0(t). \label{eq:un8}
\end{align}
In \eqref{eq:un8}, subtracting both sides from $\vl_0(t+1)$, we have
\begin{align}
& \veps(t+1) = \veps(t) + \mu \mathrm{diag}\big(\vec{\delta}(t)\big) \vl(t) \vec{\delta}^T(t)\veps(t) + \mu \mathrm{diag}\big(\vec{\delta}(t)\big) \vl(t) e_0(t) \nn \\ & - 2 \mu \mathrm{diag}\big(\vec{\delta}(t)\big) \vl_1(t) \vec{\delta}^T(t)\veps(t) - 2 \mu \mathrm{diag}\big(\vec{\delta}(t)\big) \vl_1(t) e_0(t) \nn \\ & + \big[\vl_0(t+1) - \vl_0(t)\big]. \label{eq:un9}
\end{align}
Taking expectation of both sides of \eqref{eq:un9} and using
\begin{align}
& E\big[\mu \mathrm{diag}\big(\vec{\delta}(t)\big) \vl(t) e_0(t)\big] = E\big[\mu \mathrm{diag}\big(\vec{\delta}(t)\big) \vl(t) \big]E[e_0(t)] = 0, \nn \\
& E\big[2 \mu \mathrm{diag}\big(\vec{\delta}(t)\big) \vl_1(t) e_0(t)\big] = E\big[2 \mu \mathrm{diag}\big(\vec{\delta}(t)\big) \vl_1(t) \big]E[e_0(t)] = 0, \nn
\end{align}
and assuming that $\vl_1(t)$ and $\vl_2(t)$ are independent of $\veps(t)$ \cite{transientaffine} yield
\begin{align}
& E\big[\veps(t+1)\big] = E\big[I - \mu \mathrm{diag}\big(\vl_1(t) + \vl_2(t) \big) \vec{\delta}(t) \vec{\delta}^T(t)\big]E\big[\veps(t)\big] \nn \\ & + E\big[\vl_0(t+1) - \vl_0(t)\big]. \label{eq:un10}
\end{align}
Assuming convergence of $\vR(t)$ and $\vp(t)$ (which is true for a wide range of adaptive methods in the first stage \cite{transientconvex},\cite{sayed,dsp_5}), we obtain $\limitt E\big[\vl_0(t+1) - \vl_0(t)\big] = 0$. If $\mu$ is chosen such that the eigenvalues of $E\big[I - \mu \mathrm{diag}\big(\vl_1(t) + \vl_2(t) \big) \vec{\delta}(t) \vec{\delta}^T(t)\big]$ have strictly less than unit magnitude for every $t$, then \\ $\limitt E\big[\vl(t)\big] = \limitt \vl_0(t)$.

For the transient analysis of the MSE, we have
\begin{align}
E[e^2(t)] & = E\left\{\big[y(t)-\hat{y}_m(t)\big]^2\right\}-2 \bar{\vl}_a^T(t)E\left\{\big[y(t)-\hat{y}_m(t)\big][\vec{\delta}(t);-\vec{\delta}(t)] \right\} \nn \\
          & + E\left\{ \vl_a^T(t) [\vec{\delta}(t);-\vec{\delta}(t)][\vec{\delta}(t);-\vec{\delta}(t)]^T \vl_a(t) \right\}, \nn \\
          & = E\left\{\big[y(t)-\hat{y}_m(t)\big]^2\right\}-2 \bar{\vl}_a^T(t)E\left\{\big[y(t)-\hat{y}_m(t)\big]\vu(t) \right\} \nn \\
          & + \tr \bigg( E\left[ \vl_a(t) \vl_a^T(t) \right] E \left\{ \vu(t)\vu(t)^T \right\} \bigg), \nn \\
          & = E\left\{\big[y(t)-\hat{y}_m(t)\big]^2\right\}-2 \bar{\vl}_a^T(t)\vgamma(t) + \tr \bigg( E\left[ \vl_a(t) \vl_a^T(t) \right]\vGamma(t)  \bigg), \label{eq:error}
\end{align}
where we define $\vgamma(t) \defi E\left\{\vu(t)\big[y(t)-\hat{y}_m(t)\big]\right\}$ and $\vGamma(t) \defi E\big[\vu(t)\vu^T(t)\big]$.

For the recursion of $\bar{\vl}_a(t) = E[\vl_a(t)]$, using \eqref{eq:EGU1}, we get
\begin{align}
\bar{\vl}_a(t+1) = \bar{\vl}_a(t) + \mu \mathrm{diag}\big(\vgamma(t)\big)\bar{\vl}_a(t) - \mu\mathrm{diag}\big(E[\vl_a(t)\vl_a^T(t)]\vGamma(t)\big). \label{eq:mean}
\end{align}
Using \eqref{eq:EGU}, assuming $\vl_a(t)$ is Gaussian and assuming $\lambda_a^{(i)}(t)$ and $\lambda_a^{(j)}(t)$ are independent when $i \neq j$ \cite{transientaffine}, \cite{sayed}, we get a recursion for $E\big[\vl_a(t)\vl_a^T(t)\big]$ as
\begin{align}
& E\big[\vl_a(t+1)\vl_a^T(t+1)\big] = E\big[\vl_a(t)\vl_a^T(t)\big] + \mu \mathrm{diag}\big(\vgamma(t)\big) E\big[\vl_a(t)\vl_a^T(t)\big] \nn \\ & - \mu \mathrm{diag}\big(\vGamma(t)\bar{\vl}_a(t)\big)E\big[\vl_a(t)\vl_a^T(t)\big] \nn \\ & - \mu E\big[\mathrm{diag}^2(\vu(t)) \big]\bigg(E\big[\vl_a(t)\vl_a^T(t)\big] - \bar{\vl}_a(t) \bar{\vl}_a^T(t)\bigg)\vec{1}\bar{\vl}_a^T(t) \nn \\ & - \mu \mathrm{diag}\big(\bar{\vl}_a(t)\big)\vGamma(t)\bigg(E\big[\vl_a(t)\vl_a^T(t)\big] - \bar{\vl}_a(t) \bar{\vl}_a^T(t)\bigg) \nn \\ & + \mu E\big[\vl_a(t)\vl_a^T(t)\big] \mathrm{diag}\big(\vgamma(t)\big) - \mu E\big[\vl_a(t)\vl_a^T(t)\big]\mathrm{diag}\big(\vGamma(t)\bar{\vl}_a(t)\big) \nn \\ &  - \mu \bar{\vl}_a(t)\vec{1}^T\bigg(E\big[\vl_a(t)\vl_a^T(t)\big] - \bar{\vl}_a(t) \bar{\vl}_a^T(t)\bigg)E\big[\mathrm{diag}^2(\vu(t)) \big] \nn \\ &  - \mu \bigg(E\big[\vl_a(t)\vl_a^T(t)\big] - \bar{\vl}_a(t) \bar{\vl}_a^T(t)\bigg) \vGamma(t) \mathrm{diag}\big(\bar{\vl}_a(t)\big). \label{eq:variance}
\end{align}
Defining $\vq_a(t) \defi \bar{\vl}_a(t)$ and $\vQ_a(t) \defi E\big[\vl_a(t)\vl_a^T(t)\big]$, we express \eqref{eq:mean} and \eqref{eq:variance} as a coupled recursions  in Table \ref{table:nonlin}.
\begin{table}[h]
\centering 
\caption{Time evolution of the mean and the variance of the affinely constrained mixture weights updated with the EGU algorithm}
\begin{tabular}{l}
\hline  \scriptsize
$\vq_a(t+1) = \vq_a(t) + \mu \mathrm{diag}\big(\vgamma(t)\big)\vq_a(t) - \mu\mathrm{diag}\big(\vQ_a(t)\vGamma(t)\big)$, \\ \scriptsize
$\vQ_a(t+1) = \bigg( I + \mu \mathrm{diag}\big(\vgamma(t)\big) -  \mu \mathrm{diag}\big(\vGamma(t)\vq_a(t)\big) \bigg) \vQ_a(t) - \mu E\big[\mathrm{diag}^2(\vu(t)) \big]\bigg(\vQ_a(t) - \vq_a(t)\vq_a^T(t)\bigg)\vec{1}\vq_a^T(t)$ \\ \scriptsize
$- \mu \mathrm{diag}\big(\vq_a(t)\big)\vGamma(t)\bigg( \vQ_a(t) - \vq_a(t)\vq_a^T(t)\bigg) + \vQ_a(t)  \bigg( \mu\mathrm{diag}\big(\vgamma(t)\big) - \mu\mathrm{diag}\big(\vGamma(t)\vq_a(t)\big)\bigg)$ \\ \scriptsize
$- \mu \vq_a(t)\vec{1}^T \bigg(\vQ_a(t) - \vq_a(t)\vq_a^T(t)\bigg)E\big[\mathrm{diag}^2(\vu(t)) \big] - \mu \bigg(\vQ_a(t) - \vq_a(t)\vq_a^T(t)\bigg) \vGamma(t) \mathrm{diag}\big(\vq_a(t)\big).$ \\
\hline
\end{tabular}
\label{table:nonlin}
\end{table}

In Table \ref{table:nonlin}, we provide the mean and the variance recursions for $\vQ_a(t)$ and $\vq_a(t)$. To implement these recursions, one needs to only provide $\vGamma(t)$ and $\vgamma(t)$. Note that $\vGamma(t)$ and $\vgamma(t)$ are derived for a wide range of adaptive filters \cite{transientconvex},\cite{sayed}. If we use the mean and the variance recursions in \eqref{eq:error}, then we obtain the time evolution of the final MSE. This completes the transient analysis of the affinely constrained mixture weights updated with the EGU algorithm.

\subsubsection{Relative Entropy}
For the affinely constrained combination updated with the EG algorithm, we have the multiplicative updates as \scriptsize
\begin{align}
& \lambda_1^{(i)}(t+1) = u \frac{\lambda_1^{(i)}(t) \exp\left\{  \mu e(t)  (\hat{y}_i(t)-\hat{y}_m(t)) \right\}}{\sum_{k=1}^{m-1}\bigg[ \lambda_1^{(k)}(t) \exp\left\{  \mu e(t)  (\hat{y}_k(t)-\hat{y}_m(t))\right\} + \lambda_2^{(k)}(t) \exp\left\{ -  \mu e(t)  (\hat{y}_k(t)-\hat{y}_m(t))\right\}\bigg] }, \nn \\
& \lambda_2^{(i)}(t+1) = u \frac{\lambda_2^{(i)}(t) \exp\left\{ -  \mu e(t)  (\hat{y}_i(t)-\hat{y}_m(t)) \right\}}{\sum_{k=1}^{m-1}\bigg[ \lambda_1^{(k)}(t) \exp\left\{  \mu e(t)  (\hat{y}_k(t)-\hat{y}_m(t))\right\} + \lambda_2^{(k)}(t) \exp\left\{ -  \mu e(t)  (\hat{y}_k(t)-\hat{y}_m(t))\right\}\bigg]}, \nn
\end{align} \normalsize
for $i=1,\ldots,m-1$. Using the same approximations as in \eqref{eq:un1}, \eqref{eq:un2}, \eqref{eq:un3} and \eqref{eq:un4}, we obtain \scriptsize
\begin{align}
& \lambda_1^{(i)}(t+1) = u \frac{\lambda_1^{(i)}(t) \big(1 +  \mu e(t)  (\hat{y}_i(t)-\hat{y}_m(t)) \big)}{\sum_{k=1}^{m-1}\bigg[ \lambda_1^{(k)}(t) \big(1 +  \mu e(t)  (\hat{y}_k(t)-\hat{y}_m(t)) \big) + \lambda_2^{(k)}(t) \big(1 -  \mu e(t)  (\hat{y}_k(t)-\hat{y}_m(t)) \big) \bigg]} , \label{eq:pm5} \\
& \lambda_2^{(i)}(t+1) = u \frac{\lambda_2^{(i)}(t) \big(1 -  \mu e(t)  (\hat{y}_i(t)-\hat{y}_m(t)) \big)}{\sum_{k=1}^{m-1}\bigg[ \lambda_1^{(k)}(t) \big(1 +  \mu e(t)  (\hat{y}_k(t)-\hat{y}_m(t)) \big) + \lambda_2^{(k)}(t) \big(1 -  \mu e(t)  (\hat{y}_k(t)-\hat{y}_m(t))  \bigg]}. \label{eq:pm6}
\end{align} \normalsize
In our simulations, we illustrate the accuracy of the approximations \eqref{eq:pm5} and \eqref{eq:pm6} under the mixture framework. Using \eqref{eq:pm5} and \eqref{eq:pm6}, we  obtain updates on $\vl_1(t)$ and $\vl_2(t)$ as
\begin{align}
& \vl_1(t+1) = u \frac {\big(I +  \mu e(t) \mathrm{diag}\big(\vec{\delta}(t)\big) \big)\vl_1(t) } {\big[\vec{1}^T + \mu e(t) \vu^T(t) \big]\vl_a(t)}, \label{eq:pm7} \\
& \vl_2(t+1) = u \frac {\big(I -  \mu e(t) \mathrm{diag}\big(\vec{\delta}(t)\big) \big)\vl_2(t)}{\big[\vec{1}^T + \mu e(t) \vu^T(t) \big]\vl_a(t)} . \label{eq:pm8}
\end{align}
Using updates \eqref{eq:pm7} and \eqref{eq:pm8}, we can write update on $\vl_a(t)$ 
\begin{align}
\vl_a(t+1)= u \frac{\big[I + \mu e(t) \mathrm{diag}\big(\vu(t)\big) \big]\vl_a(t)}{\big[\vec{1}^T + \mu e(t) \vu^T(t) \big]\vl_a(t)}. \label{eq:re}
\end{align}
For the recursion of $\bar{\vl}_a(t)$, using \eqref{eq:re}, we get
\begin{align}
E\big[\vl_a(t+1)\big] & = E\left\{ u \frac{\big[I + \mu e(t) \mathrm{diag}\big(\vu(t)\big) \big]\vl_a(t)}{\big[\vec{1}^T + \mu e(t) \vu^T(t) \big]\vl_a(t)} \right\}, \nn \\
                      & \approx u \frac{E\left\{\big[I + \mu e(t) \mathrm{diag}\big(\vu(t)\big) \big]\vl_a(t)\right\}}{E\left\{\big[\vec{1}^T + \mu e(t) \vu^T(t) \big]\vl_a(t)\right\}}, \label{eq:assume} \\
                      & = u \frac{E\big[\vl_a(t)\big] + \mu \mathrm{diag}\big(\vgamma(t)\big)E\big[\vl_a(t)\big] - \mu\mathrm{diag}\big(E[\vl_a(t)\vl_a^T(t)]\vGamma(t)\big)}{\big[\vec{1}^T + \mu \vgamma^T(t) \big]E\big[\vl_a(t)\big] - \mu \tr\big(E[\vl_a(t)\vl_a^T(t)]\vGamma(t)\big)}, \label{eq:123}
\end{align}
where in \eqref{eq:assume} we approximate expectation of the quotient with the quotient of the expectations. In our simulations, we also illustrate the accuracy of this approximation in the mixture framework. From \eqref{eq:re}, using the same approximation in \eqref{eq:123}, assuming $\vl_a(t)$ is Gaussian, assuming $\lambda_a^{(i)}(t)$ and $\lambda_a^{(j)}(t)$ are independent when $i \neq j$, we get a recursion for $E\big[\vl_a(t)\vl_a^T(t)\big]$ as
\begin{align}
E\big[\vl_a(t+1)\vl_a^T(t+1)\big] = u^2 \frac{\vec{A}(t)}{b(t)}, \label{eq:pm10}
\end{align}
where $\vec{A}(t)$ is equal to the right hand side of \eqref{eq:variance} and
\begin{align}
& b(t) = \vec{1}^T E\big[\vl_a(t)\vl_a^T(t)\big] \vec{1} + \mu \vp^T(t) E\big[\vl_a(t)\vl_a^T(t)\big] \vec{1} \nn \\ & - \mu\bar{\vl}_a^T(t)\vR(t)E\big[\vl_a(t)\vl_a^T(t)\big]\vec{1} - \mu\vec{1}^T\bigg(E\big[\vl_a(t)\vl_a^T(t)\big] - \bar{\vl}_a(t) \bar{\vl}_a^T(t)\bigg)\vR(t)\bar{\vl}_a(t) \nn \\ & - \mu\vec{1}^T\bigg(E\big[\vl_a(t)\vl_a^T(t)\big] - \bar{\vl}_a(t) \bar{\vl}_a^T(t)\bigg)E\big[\mathrm{diag}^2(\vu(t)) \big]\vec{1}^T\bar{\vl}_a(t)\vec{1} \nn \\ & + \mu\vec{1}^TE\big[\vl_a(t)\vl_a^T(t)\big]\vp(t) - \mu\vec{1}^TE\big[\vl_a(t)\vl_a^T(t)\big]\vR(t)\bar{\vl}_a(t) \nn \\ & -  \mu\bar{\vl}_a^T(t)\vR(t)\bigg(E\big[\vl_a(t)\vl_a^T(t)\big] - \bar{\vl}_a(t) \bar{\vl}_a^T(t)\bigg)\vec{1} \nn \\ & - \mu\vec{1}^T\bar{\vl}_a^T(t)\vec{1}E\big[\mathrm{diag}^2(\vu(t)) \big]\bigg(E\big[\vl_a(t)\vl_a^T(t)\big] - \bar{\vl}_a(t) \bar{\vl}_a^T(t)\bigg)\vec{1}. \label{eq:pm11}
\end{align}
If we use the mean \eqref{eq:123} and the variance \eqref{eq:pm10}, \eqref{eq:pm11} recursions in \eqref{eq:error}, then we obtain the time evolution of the final MSE. This completes the transient analysis of the affinely constrained mixture weights updated with the EG algorithm.
\subsection{Unconstrained Mixture}
We use the unconstrained relative entropy and the relative entropy as distance measures to update unconstrained mixture weights. We first perform transient analysis of the mixture weights updated using the EGU algorithm. Then, we continue with the transient analysis of the mixture weights updated using the EG algorithm. Note that since the unconstrained case is close to the affinely constrained case, we only provide the necessary modifications to get the mean and the variance recursions for the transient analysis.
\subsubsection{Unconstrained Relative Entropy}
For the unconstrained combination updated with EGU, we have the multiplicative updates as
\begin{align}
w_1^{(i)}(t+1) & = w_1^{(i)}(t) \exp\left\{ \mu e(t) \hat{y}_i(t) \right\}, \nn \\
w_2^{(i)}(t+1) & = w_2^{(i)}(t) \exp\left\{- \mu e(t) \hat{y}_i(t) \right\}, \nn
\end{align}
for $i=1,\ldots,m$. Using the same approximations as in \eqref{eq:un1}, \eqref{eq:un2}, \eqref{eq:un3} and \eqref{eq:un4}, we can obtain updates on $\vw_1(t)$ and $\vw_2(t)$  as
\begin{align}
& \vw_1(t+1) =  \big(I +  \mu e(t) \mathrm{diag}\big(\vx(t)\big) \big)\vw_1(t) , \label{eq:unc1} \\
& \vw_2(t+1) =  \big(I -  \mu e(t) \mathrm{diag}\big(\vx(t)\big) \big)\vw_2(t). \label{eq:unc2}
\end{align}
Collecting the weights in $\vw_a(t) = [\vw_1(t);\vw_2(t)]$, using the updates \eqref{eq:unc1} and \eqref{eq:unc2}, we can write update on $\vw_a(t)$ as
\begin{align}
\vw_a(t+1)= \big(I + \mu e(t) \mathrm{diag}\big(\vu(t)\big) \big)\vw_a(t), \label{eq:EGU}
\end{align}
where $\vu(t)$ is defined as $\vu(t)\defi [\vx(t);-\vx(t)]$.

For the desired signal $y(t)$, we can write $ y(t) = \vw_0^T(t) \vx(t) + e_0(t)$, where $\vw_0(t)$ is the optimum MSE solution at time $t$ such that $\vw_0(t) \defi \vR^{-1}(t)\vp(t)$, $\vR(t) \defi E\big[\vx(t)\vx^T(t)\big]$, $\vp(t) \defi E\left\{\vx(t)y(t)\right\}$ and $e_0(t)$ is zero-mean disturbance uncorrelated to $\vx(t)$. To show that the mixture weights converge to the optimum solution in the steady-state such that $\limitt E\big[\vw(t)\big] = \limitt \vw_0(t)$, we follow similar lines as in the Section 4.1.1. We modify \eqref{eq:un7}, \eqref{eq:un8}, \eqref{eq:un9} and \eqref{eq:un10} such that $\vl$ will be replaced by $\vw$, $\vec{\delta}(t)$ will be replaced by $\vx(t)$ and $\veps(t) = \vw_0(t) - \vw(t)$. After these replacements, we obtain
\begin{align}
& E\big[\veps(t+1)\big] = E\big[I - \mu \mathrm{diag}\big(\vw_1(t) + \vw_2(t) \big) \vx(t) \vx^T(t)\big]E\big[\veps(t)\big] \nn \\ & + E\big[\vw_0(t+1) - \vw_0(t)\big]. \label{eq:unc10}
\end{align}
Since, we have $\limitt E\big[\vw_0(t+1) - \vw_0(t)\big] = 0$ for most adaptive filters in the first stage \cite{sayed} and if $\mu$ is chosen so that all the eigenvalues of $E\big[I - \mu \mathrm{diag}\big(\vw_1(t) + \vw_2(t) \big) \vx(t) \vx^T(t)\big]$ have strictly less than unit magnitude for every $t$, then $\limitt E\big[\vw(t)\big] = \limitt \vw_0(t)$.

For the transient analysis of MSE, defining $\vgamma(t) \defi E\left\{\vu(t)y(t)\right\}$ and $\vGamma(t) \defi E\big[\vu(t)\vu^T(t)\big]$, \eqref{eq:error} is modified as
\begin{align}
E[e^2(t)] & = E\left\{y^2(t)\right\}-2 \bar{\vw}_a^T(t)\vgamma(t) + \tr \bigg( E\left[ \vw_a(t) \vw_a^T(t) \right]\vGamma(t)  \bigg). \label{eq:error2}
\end{align}
Accordingly, we modify the mean recursion \eqref{eq:mean} and the variance recursion \eqref{eq:variance} such that instead of $\vl_a(t)$ we use $\vw_a(t)$. We also modify the Table \ref{table:nonlin} using $\vq_a(t) \defi \bar{\vw}_a(t)$ and $\vQ_a(t) \defi E\big[\vw_a(t)\vw_a^T(t)\big]$. If we use this modified mean and variance recursions in \eqref{eq:error2}, then we obtain the time evolution of the final MSE. This completes the transient analysis of the unconstrained mixture weights updated with the EGU algorithm.

\subsubsection{Relative Entropy}
For the unconstrained combination updated with the EG algorithm, we have the multiplicative updates as
\begin{align}
& w_a^{(i)}(t+1) = u \frac{w_a^{(i)}(t) \exp\left\{  \mu e(t)  \hat{y}_i(t) \right\}}{\displaystyle \sum_{k=1}^m \bigg[ w_a^{(k)}(t) \exp\left\{  \mu e(t)  \hat{y}_k(t) \right\} + w_a^{(k+m)}(t) \exp\left\{ - \mu e(t)  \hat{y}_k(t) \right\} \bigg]}, \nn \\
& i=1,\ldots,m, \nn \\
& w_a^{(i)}(t+1) = u \frac{w_a^{(i)}(t) \exp\left\{ -  \mu e(t)  \hat{y}_i(t) \right\}}{\displaystyle \sum_{k=1}^m \bigg[ w_a^{(k)}(t) \exp\left\{  \mu e(t)  \hat{y}_k(t) \right\} + w_a^{(k+m)}(t) \exp\left\{ - \mu e(t)  \hat{y}_k(t) \right\} \bigg]}, \nn \\
& i=m+1\ldots,2m. \nn
\end{align}
Following similar lines, we modify \eqref{eq:pm7}, \eqref{eq:pm8},
\eqref{eq:re}, \eqref{eq:123}, \eqref{eq:pm10} and \eqref{eq:pm11}
such that we replace $\vec{\delta}(t)$ with $\vx(t)$, $\vl$ with $\vw$
and $\vu(t) = \big[\vx(t);-\vx(t)\big]$. Finally, we use the modified
mean and variance recursions in \eqref{eq:error2} and obtain the time
evolution of the final MSE. This completes the transient analysis of
the unconstrained mixture weights updated with the EG algorithm.

\section{Simulations}
In this section, we illustrate the accuracy of our results and compare
performances of different adaptive mixture methods through
simulations. In our simulations, we observe that using the EG
algorithm to train the mixture weights yields better performance
compared to using the LMS algorithm or the EGU algorithm to train the mixture
weights for combinations having more than two filters and when the
combination favors only a few of the constituent filters. The LMS algorithm and
the EGU algorithm perform similarly in our simulations when they are used
to train the mixture weights. We also observe in our simulations that
the mixture weights under the EG update converge to the optimum
combination vector faster than the mixture weights under the LMS
algorithm.

\begin{figure}[t]
\centerline{\epsfxsize=8cm \epsfbox{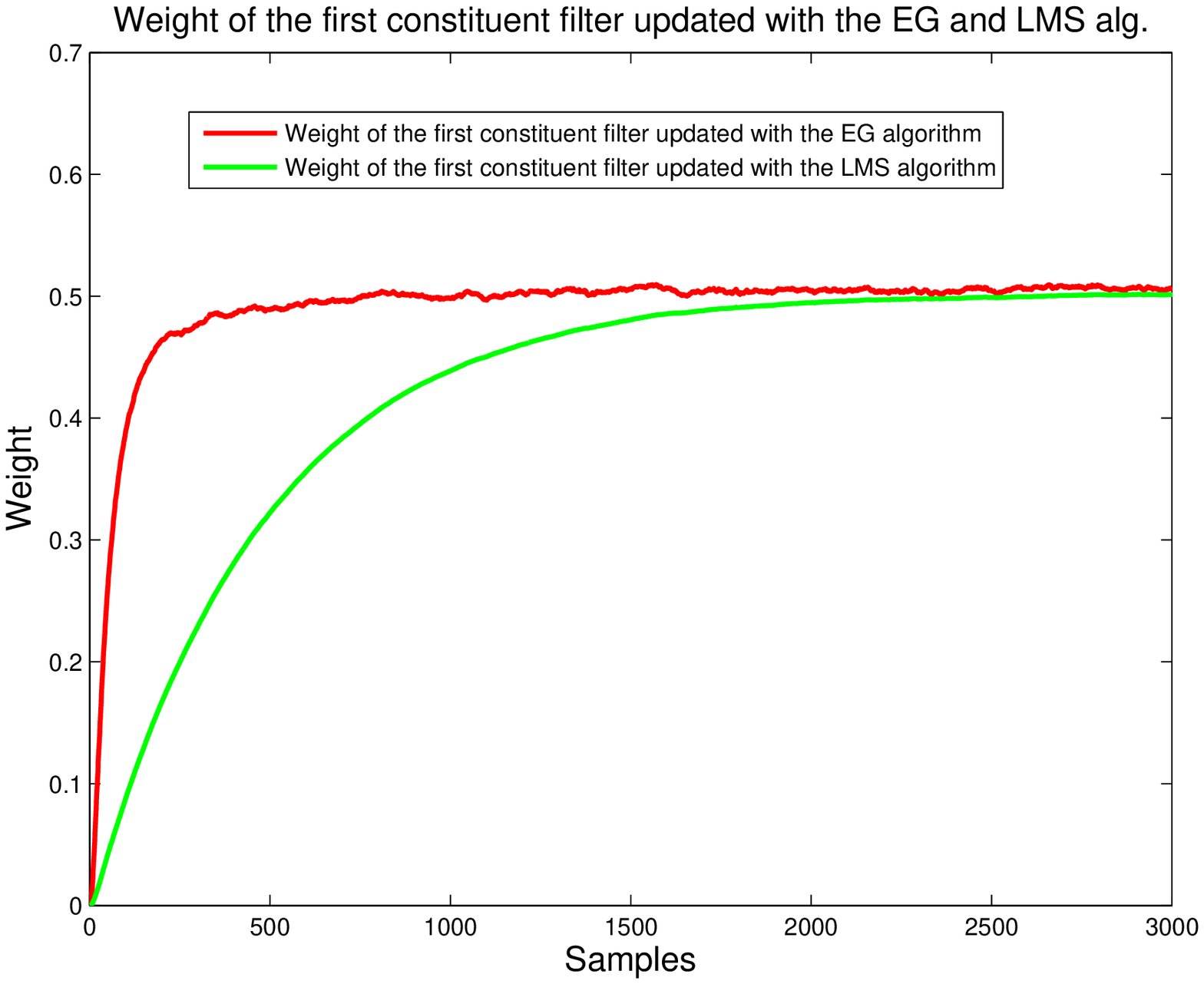}
\hspace{0.2in} \epsfxsize=8cm \epsfbox{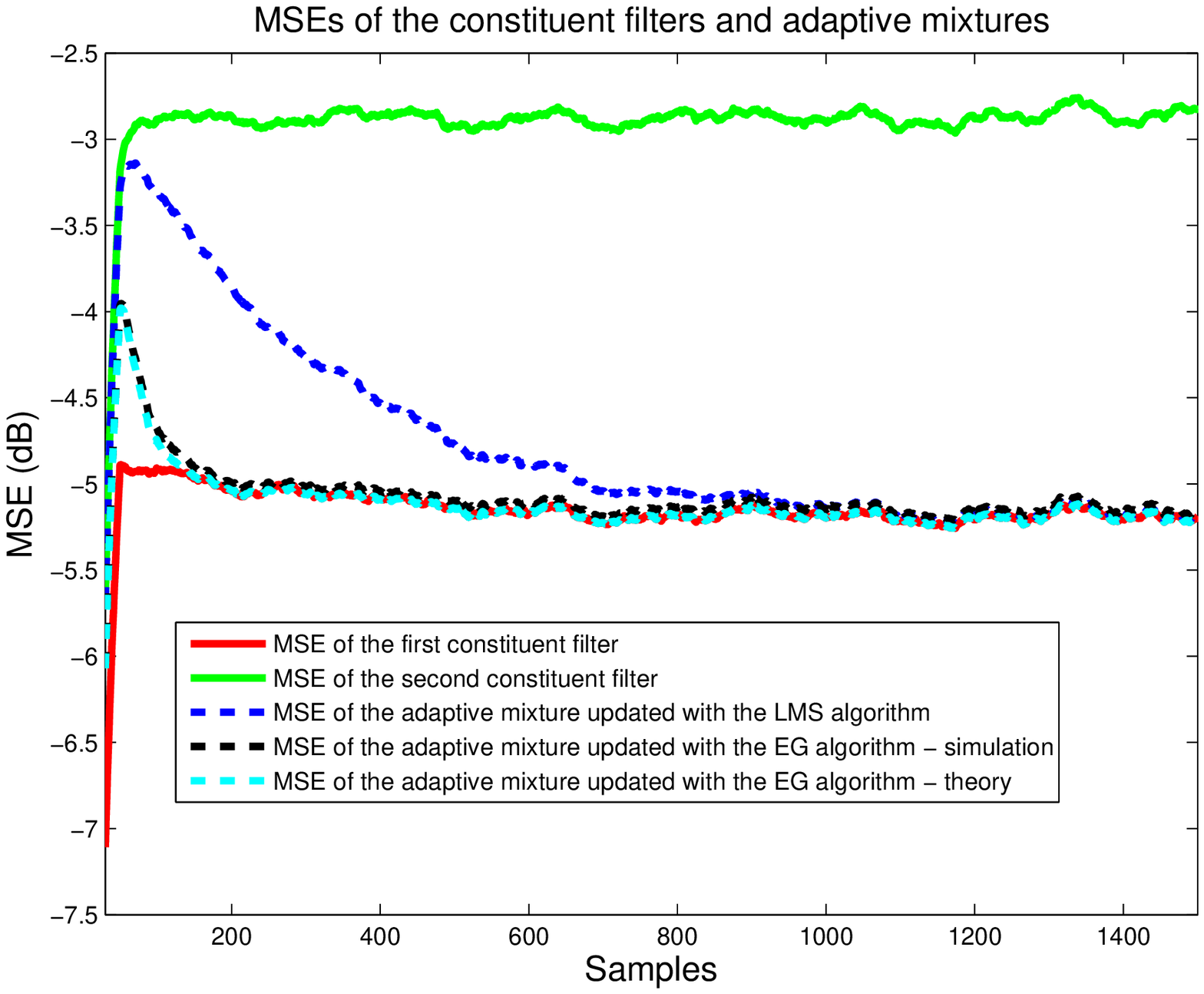}}
\centerline{\hspace{1in}(a)\hspace{3.3in}(b)\hfill}
\centerline{\epsfxsize=8cm \epsfbox{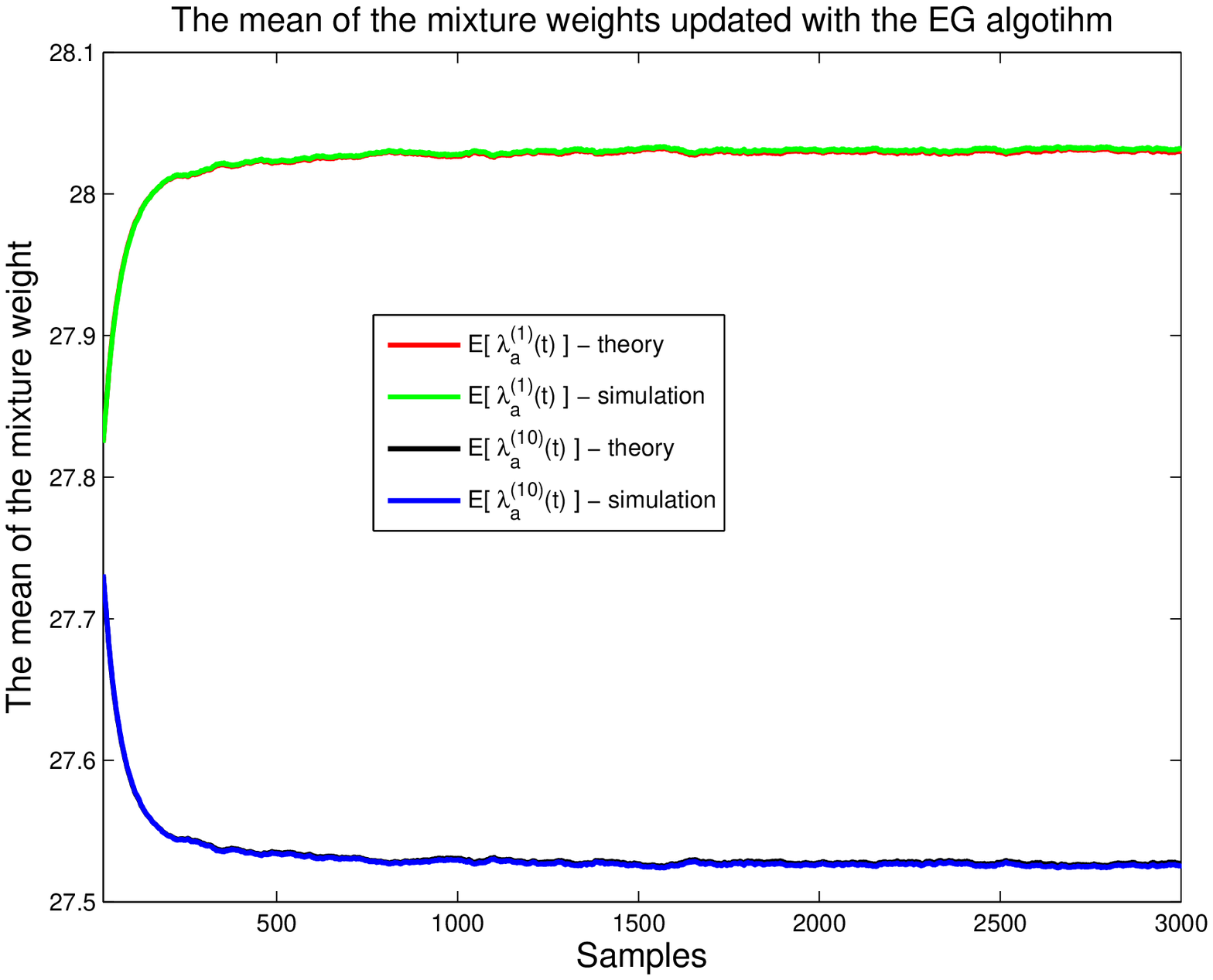}
\hspace{0.2in} \epsfxsize=8cm \epsfbox{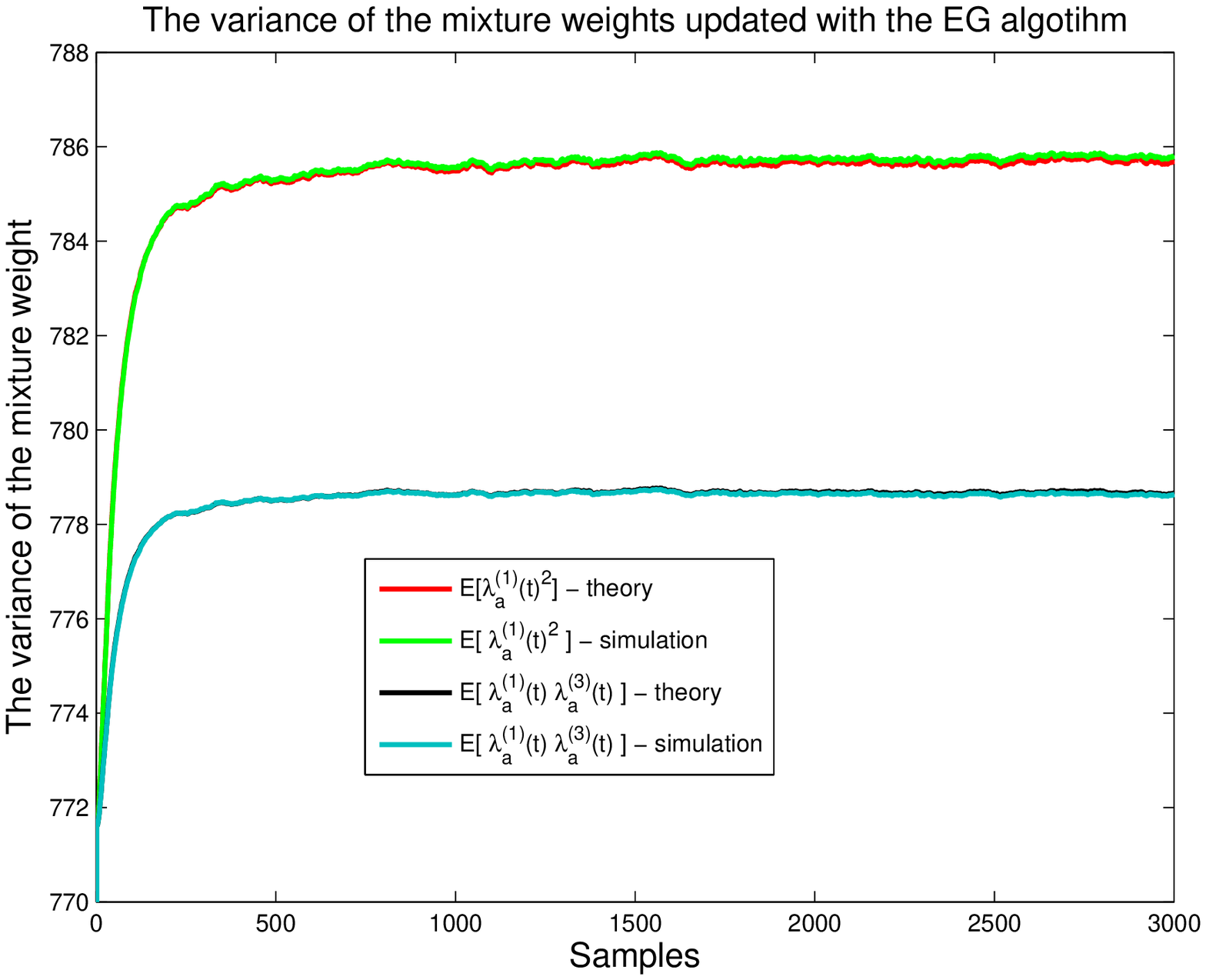}}
\centerline{\hspace{1in}(c)\hspace{3.3in}(d)\hfill}
\caption{ Using 10 LMS filters as constituent filters, where learning rates for 2 constituent filters are $\mu = 0.002$ and for the rest are $\mu \in [0.1,0.11]$. SNR = -10dB. For the mixture stage, the EG algorithm has $\mu_{\mathrm{EG}} = 0.0008$ and the LMS algorithm has $\mu_{\mathrm{LMS}} = 0.005$. For the EG algorithm, $u = 500$. (a) The weight of the first constituent filter in the mixture, i.e., $E[\vl^{(1)}(t)]$. (b) The MSE curves for adaptive mixture updated with the EG algorithm, the adaptive mixture updated with the LMS algorithm, the first constituent filter and the second constituent filter. (c) Theoretical values $\bar{\lambda}^{(1)}_a(t)$ and $\bar{\lambda}^{(10)}_a(t)$ and simulations. (d) Theoretical values $E\big[\lambda_a^{(1)}(t)^2\big]$ and $E\big[\lambda_a^{(1)}(t)\lambda_a^{(3)}(t)\big]$ and simulations.
\label{fig:1}}
\end{figure}

To compare performances of the EG and LMS algorithms and illustrate
the accuracy of our results in \eqref{eq:123}, \eqref{eq:pm10} and
\eqref{eq:pm11} under different algorithmic parameters, the desired
signal as well as the system parameters are selected as
follows. First, a seventh-order linear filter, \\ $\vw_o = [0.25,
  -0.47, -0.37, 0.045, -0.18, 0.78, 0.147]^T$, is chosen as in 
\cite{transientaffine}. The underlying signal is generated using the
data model $y(t) = \tau \: \vw_o^T\va(t)+n(t)$, where $\va(t)$ is an
i.i.d. Gaussian vector process with zero mean and unit variance
entries, i.e., $E[\va(t) \va^T(t)]=\vec{I}$, $n(t)$ is an i.i.d.
Gaussian noise process with zero mean and variance $E[n^2(t)]=0.3$,
and $\tau$ is a positive scalar to control SNR. Hence, the SNR of the
desired signal is given by $\mbox{SNR} \defi
10\log(\frac{E[\tau^2(\vw_{o}^T\vu(t))^2]}{0.01})=10\log(\frac{\tau^2
  \|\vw_o\|^2}{0.01})$. For the first experiment, we have SNR =
-10dB. To model the unknown system we use ten linear filters using the
LMS update as the constituent filters. The learning rates of these two
constituent filters are set to $\mu_1 = 0.002$ and $\mu_6 = 0.002$
while the learning rates for the rest of the constituent filters are
selected randomly in $[0.1,0.11]$. Therefore, in the steady-state, we
obtain the optimum combination vector approximately as $\vl_o = [0.5,
  0, 0, 0, 0, 0.5, 0, 0, 0, 0]^T$, i.e., the final combination vector
is sparse. In the second stage, we train the combination weights with
the EG and LMS algorithms and compare performances of these
algorithms. For the second stage, the learning rates for the EG and
LMS algorithms are selected as $\mu_{\mathrm{EG}} = 0.0008$ and
$\mu_{\mathrm{LMS}} = 0.005$ such that the MSEs of both mixtures
converge to the same final MSE to provide a fair comparison. We select
$u = 500$ for the EG algorithm.  In Fig. \ref{fig:1}a, we plot the
weight of the first constituent filter with $\mu_1 = 0.002$,
i.e. $E[\vl^{(1)}(t)]$, updated with the EG and LMS algorithms. In
Fig. \ref{fig:1}b, we plot the MSE curves for the adaptive mixture
updated with the EG algorithm, the adaptive mixture updated with the
LMS algorithm, the first constituent filter with $\mu_1 = 0.002$ and
the second constituent filter with $\mu_2 \in [0.1,0.11]$.  From
Fig. \ref{fig:1}a and Fig. \ref{fig:1}b, we see that the EG algorithm
performs better than the LMS algorithm such that the combination
weight under the update of the EG algorithm converges to 0.5 faster
than the combination weight under the update of the LMS
algorithm. Furthermore the MSE of the adaptive mixture updated with
the EG algorithm converges faster than the MSE of the adaptive mixture
updated with the LMS algorithm. In Fig. \ref{fig:1}c, to test the
accuracy of \eqref{eq:123}, we plot the theoretical values for
$\bar{\lambda}^{(1)}_a(t)$ and $\bar{\lambda}^{(10)}_a(t)$ along with
simulations. Note in Fig. \ref{fig:1}c we observe that
$\bar{\lambda}^{(1)}(t) = \bar{\lambda}^{(1)}_a(t) -
\bar{\lambda}^{(10)}_a(t)$ converges to 0.5 as predicted in our
derivations.  In Fig. \ref{fig:1}d, to test the accuracy of
\eqref{eq:pm10} and \eqref{eq:pm11}, as an example, we plot the
theoretical values of $E\big[\lambda_a^{(1)}(t)^2\big]$ and
$E\big[\lambda_a^{(1)}(t)\lambda_a^{(3)}(t)\big]$ along with
simulations.  As we observe from Fig. \ref{fig:1}c and
Fig. \ref{fig:1}d, there is a close agreement between our results and
simulations in these experiments. We observe similar results for the
other cross terms.

\begin{figure}[t]
\centerline{\epsfxsize=8cm \epsfbox{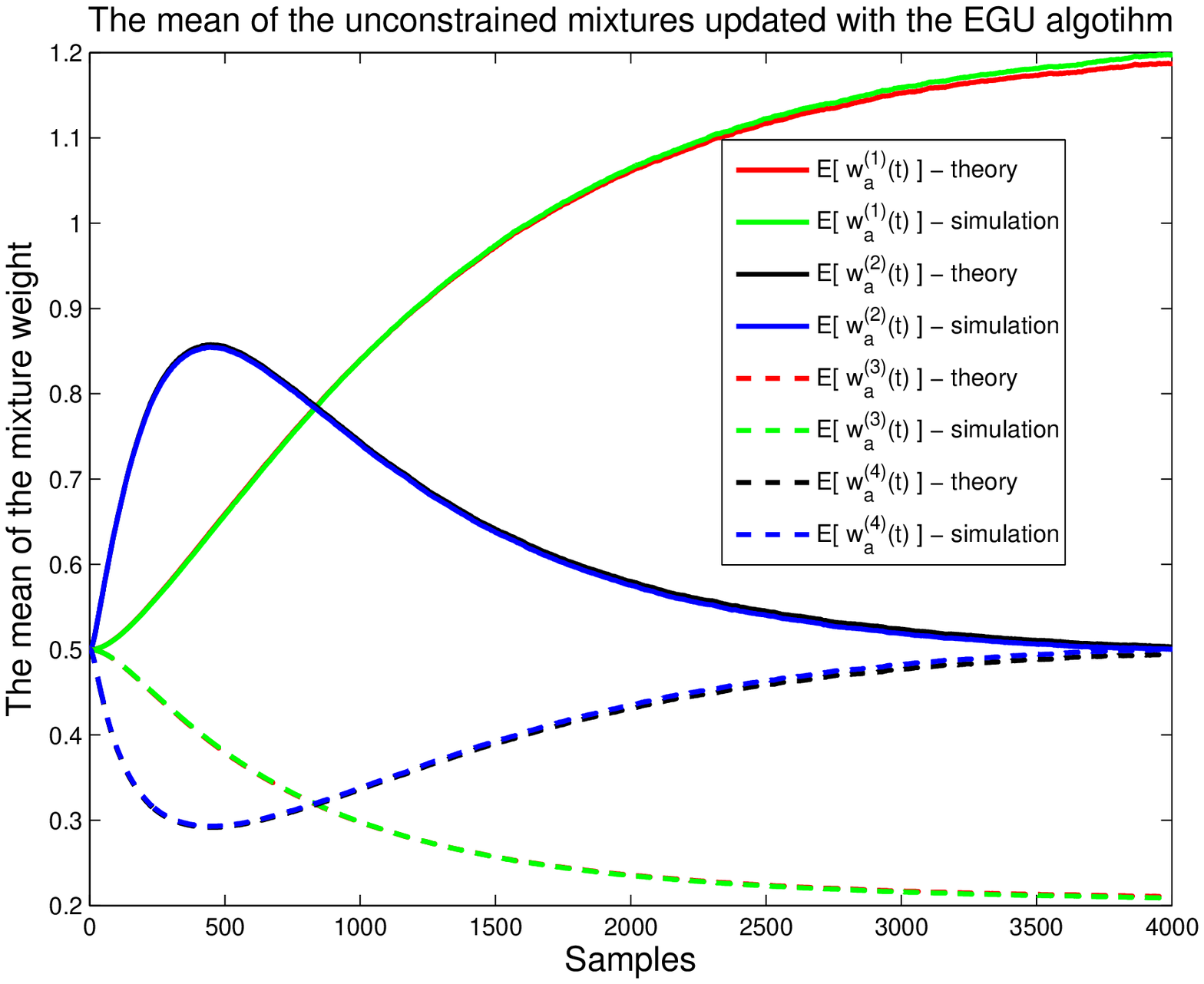}
\hspace{0.2in} \epsfxsize=8cm \epsfbox{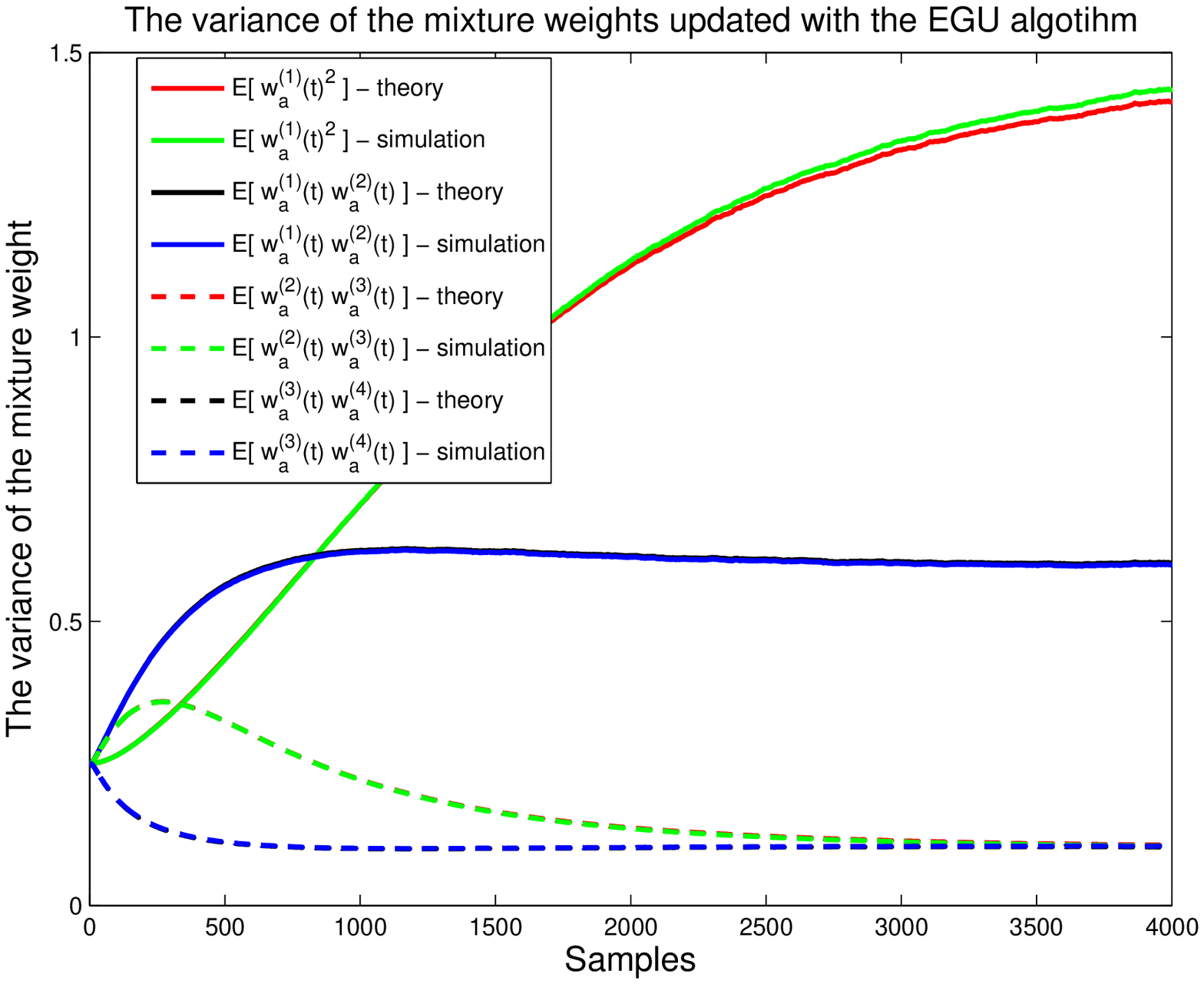} }
\centerline{\hspace{0.95in}(a)\hspace{3.25in}(b)\hfill}
\centerline{\epsfxsize=8cm \epsfbox{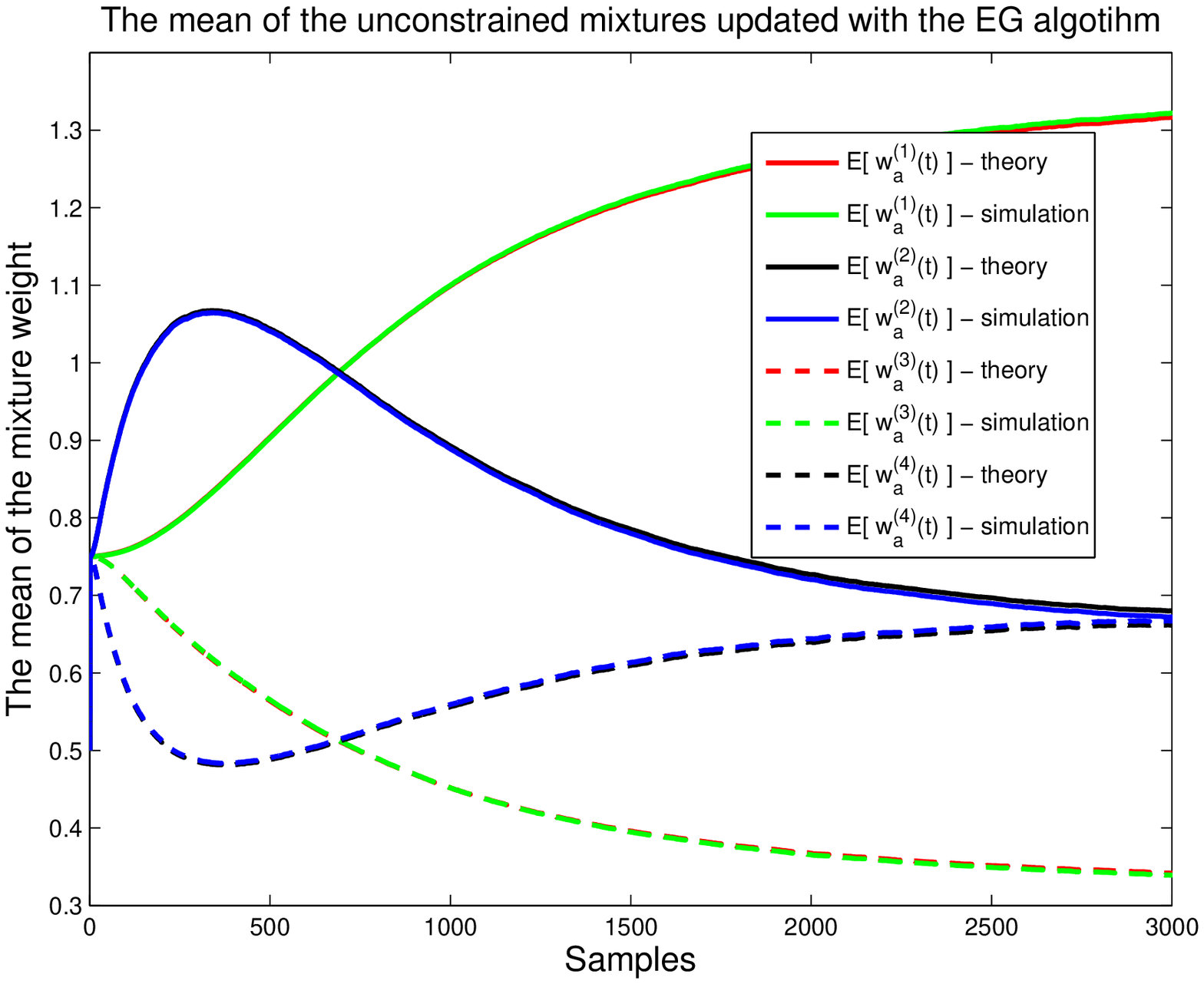}
\hspace{0.2in} \epsfxsize=8cm \epsfbox{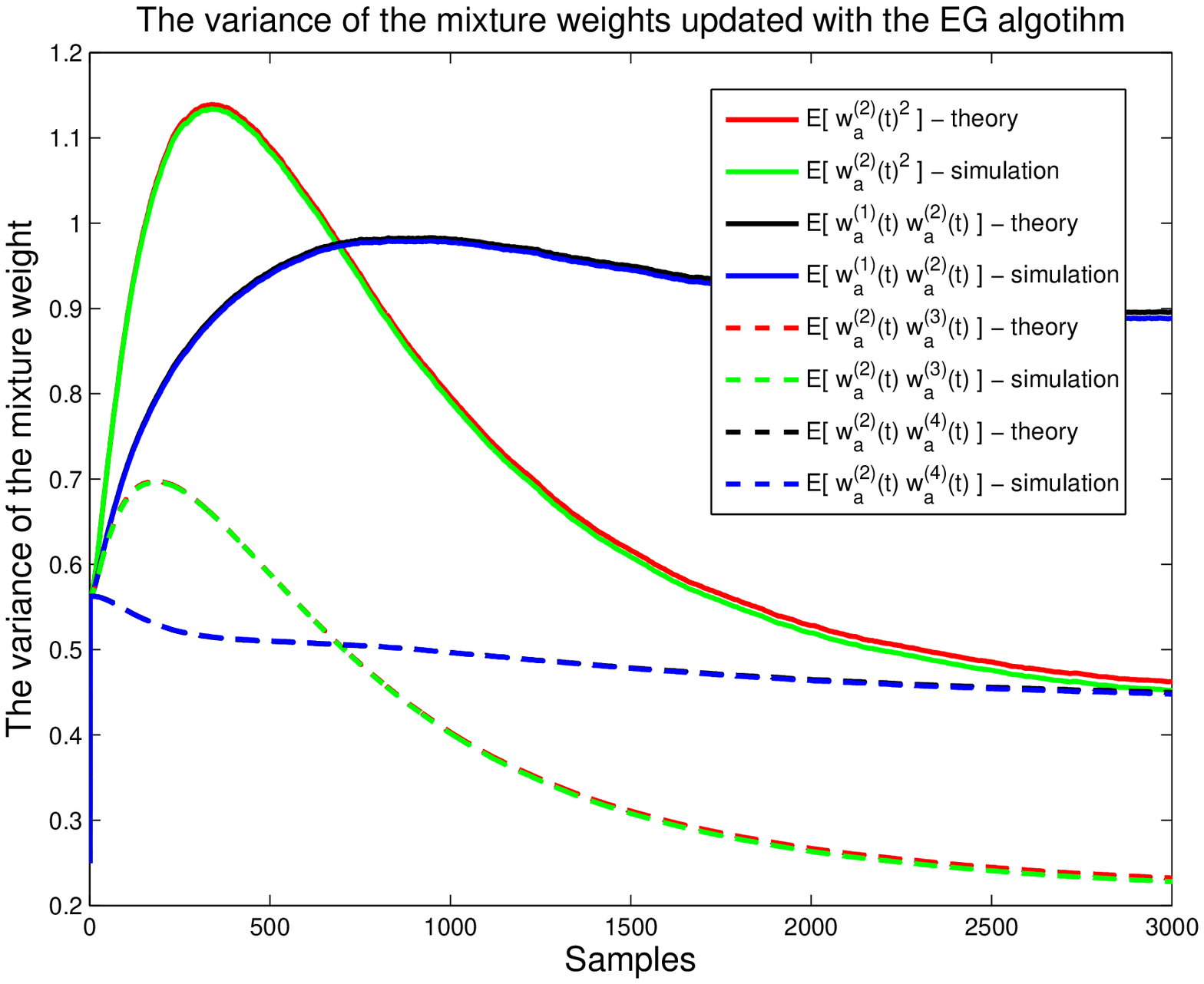} }
\centerline{\hspace{0.95in}(c)\hspace{3.25in}(d)\hfill}
\caption{ Two LMS filters as constituent filters with learning rates $\mu_1 = 0.002$ and $\mu_2 = 0.1$, respectively. SNR = 1dB. For the second stage, the EGU algorithm has $\mu_{\mathrm{EGU}} = 0.01$ and the EG algorithm has $\mu_{\mathrm{EG}} = 0.01$. For the EG algorithm, $u = 3$. (a) Theoretical values for the mixture weights updated with the EGU algorithm and simulations. (b) Theoretical values $E\big[\vw_a^{(1)}(t)^2\big]$,  $E\big[\vw_a^{(1)}(t)\vw_a^{(2)}(t)\big]$, $E\big[\vw_a^{(2)}(t)\vw_a^{(3)}(t)\big]$ and $E\big[\vw_a^{(3)}(t)\vw_a^{(4)}(t)\big]$ and simulations. (c) Theoretical mixture weights updated with the EG algorithm and simulations. (d) Theoretical values $E\big[\vw_a^{(2)}(t)^2\big]$, $E\big[\vw_a^{(1)}(t)\vw_a^{(2)}(t)\big]$, $E\big[\vw_a^{(2)}(t)\vw_a^{(3)}(t)\big]$ and $E\big[\vw_a^{(2)}(t)\vw_a^{(4)}(t)\big]$ and simulations.
\label{fig:2}}
\end{figure}

We next simulate the unconstrained mixtures updated with the EGU and EG algorithms. Here, we have two linear filters and both using the LMS update to train their weight vectors as the constituent filters. The learning rates for two constituent filters are set to $\mu_1 = 0.002$ and $\mu_2 = 0.1$ respectively. Therefore, in the steady-state, we obtain the optimum vector approximately as $\vw_o = [1, 0]$. We have SNR = 1 for these simulations. The unconstrained mixture weights are first updated with the EGU algorithm. For the second stage, the learning rate for the EGU algorithm is selected as $\mu_{\mathrm{EGU}} = 0.01$. The theoretical curves in the figures are produced using $\vGamma(t)$ and $\vgamma(t)$ that are calculated from the simulations, since our goal is to illustrate the validity of derived equations. In Fig. \ref{fig:2}a, we plot the theoretical values of $\bar{\vw}^{(1)}_a(t)$, $\bar{\vw}^{(2)}_a(t)$, $\bar{\vw}^{(3)}_a(t)$ and $\bar{\vw}^{(4)}_a(t)$ along with simulations. In Fig. \ref{fig:2}b, as an example, we plot the theoretical values of $E\big[\vw_a^{(1)}(t)^2\big]$,  $E\big[\vw_a^{(1)}(t)\vw_a^{(2)}(t)\big]$, $E\big[\vw_a^{(2)}(t)\vw_a^{(3)}(t)\big]$ and $E\big[\vw_a^{(3)}(t)\vw_a^{(4)}(t)\big]$ along with simulations. We continue to update the mixture weights with the EG algorithm. For the second stage, the learning rate for the EG algorithm is selected as $\mu_{\mathrm{EG}} = 0.01$. We select $u = 3$ for the EG algorithm. In Fig. \ref{fig:2}c, we plot the theoretical values of $\bar{\vw}^{(1)}_a(t)$, $\bar{\vw}^{(2)}_a(t)$, $\bar{\vw}^{(3)}_a(t)$ and $\bar{\vw}^{(4)}_a(t)$ along with simulations. In Fig. \ref{fig:2}d, as an example, we plot the theoretical values of $E\big[\vw_a^{(2)}(t)^2\big]$, $E\big[\vw_a^{(1)}(t)\vw_a^{(2)}(t)\big]$, $E\big[\vw_a^{(2)}(t)\vw_a^{(3)}(t)\big]$ and $E\big[\vw_a^{(2)}(t)\vw_a^{(4)}(t)\big]$ along with simulations. We observe a close agreement between our results and simulations.

\begin{figure}[t]
\centerline{\epsfxsize=8cm \epsfbox{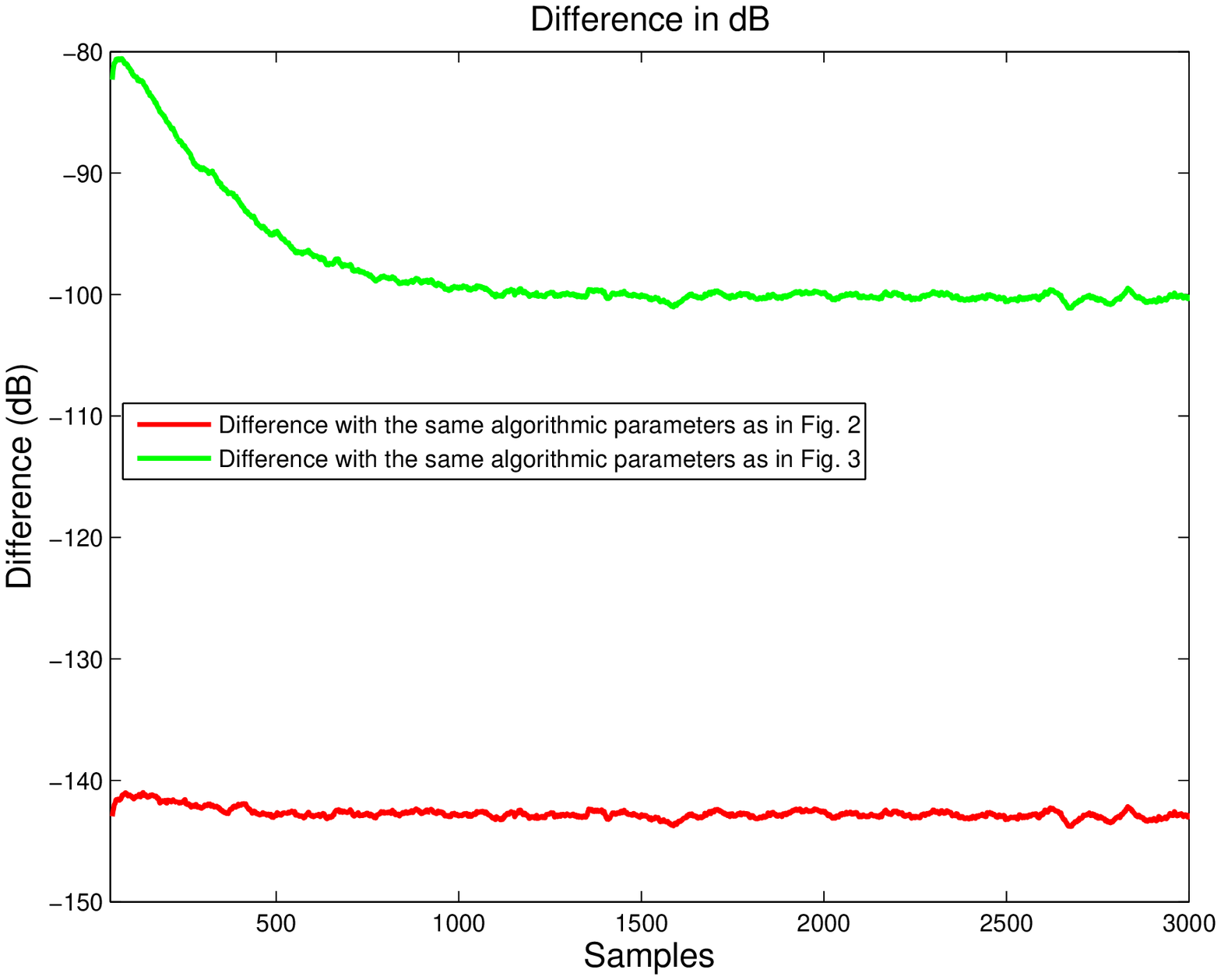}
\hspace{0.2in} \epsfxsize=8cm \epsfbox{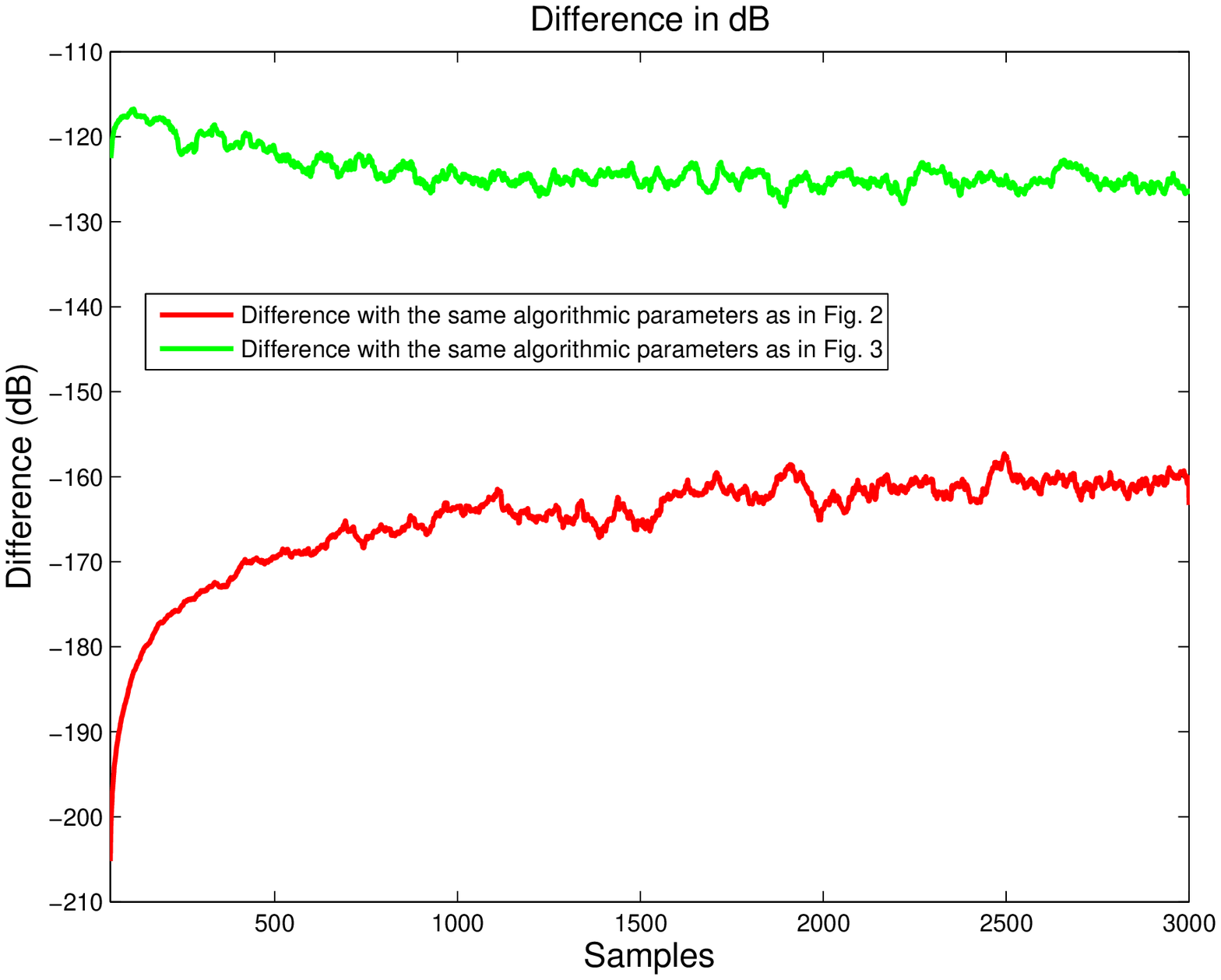} }
\centerline{\hspace{0.98in}(a)\hspace{3.27in}(b)\hfill}
\caption{ (a) The difference $\frac { \| \exp\left\{  \mu e(t)  (\hat{y}_i(t)-\hat{y}_m(t)) \right\} - \left\{1 +  \mu e(t)  (\hat{y}_i(t)-\hat{y}_m(t)) ) \right\} \|^2  } {\sqrt{\| \exp\left\{  \mu e(t)  (\hat{y}_i(t)-\hat{y}_m(t)) \right\} \|^2 \| \left\{1 +  \mu e(t)  (\hat{y}_i(t)-\hat{y}_m(t)) ) \right\} \|^2  } }$ for $i = 1$ with the same algorithmic parameters as in Fig. \ref{fig:1} and Fig. \ref{fig:2}. (b) The first parameter of the difference { $ \frac { \bigg\Vert E\left\{ u \frac{\big[I + \mu e(t) \mathrm{diag}\big(\vu(t)\big) \big]\vl_a(t)}{\big[\vec{1}^T + \mu e(t) \vu^T(t) \big]\vl_a(t)} \right\} - u \frac{E\left\{\big[I + \mu e(t) \mathrm{diag}\big(\vu(t)\big) \big]\vl_a(t)\right\}}{E\left\{\big[\vec{1}^T + \mu e(t) \vu^T(t) \big]\vl_a(t)\right\}} \bigg\Vert^2 } { \sqrt{ \bigg\Vert E\left\{ u \frac{\big[I + \mu e(t) \mathrm{diag}\big(\vu(t)\big) \big]\vl_a(t)}{\big[\vec{1}^T + \mu e(t) \vu^T(t) \big]\vl_a(t)} \right\} \bigg\Vert^2  \bigg\Vert u \frac{E\left\{\big[I + \mu e(t) \mathrm{diag}\big(\vu(t)\big) \big]\vl_a(t)\right\}}{E\left\{\big[\vec{1}^T + \mu e(t) \vu^T(t) \big]\vl_a(t)\right\}} \bigg\Vert^2 } } $ } \normalsize with the same algorithmic parameters as in Fig. \ref{fig:1} and Fig. \ref{fig:2}. \label{fig:3}}
\end{figure}

To test the accurateness of the assumptions in \eqref{eq:un3} and \eqref{eq:un4}, we plot in Fig.
\ref{fig:3}a, the difference  \\
\[
\frac { \| \exp\left\{  \mu e(t)  (\hat{y}_i(t)-\hat{y}_m(t)) \right\} - \left\{1 +  \mu e(t)  (\hat{y}_i(t)-\hat{y}_m(t)) ) \right\} \|^2  } {\sqrt{\| \exp\left\{  \mu e(t)  (\hat{y}_i(t)-\hat{y}_m(t)) \right\} \|^2 \| \left\{1 +  \mu e(t)  (\hat{y}_i(t)-\hat{y}_m(t)) ) \right\} \|^2  } }
\]
for $i = 1$ with the same algorithmic parameters as in Fig. \ref{fig:1} and Fig. \ref{fig:2}. To test the accurateness of the separation assumption in \eqref{eq:123}, we plot in Fig.
\ref{fig:3}b, the first parameter of the difference \\ \small
\[
\frac { \bigg\Vert E\left\{ u \frac{\big[I + \mu e(t) \mathrm{diag}\big(\vu(t)\big) \big]\vl_a(t)}{\big[\vec{1}^T + \mu e(t) \vu^T(t) \big]\vl_a(t)} \right\} - u \frac{E\left\{\big[I + \mu e(t) \mathrm{diag}\big(\vu(t)\big) \big]\vl_a(t)\right\}}{E\left\{\big[\vec{1}^T + \mu e(t) \vu^T(t) \big]\vl_a(t)\right\}} \bigg\Vert^2 } { \sqrt{ \bigg\Vert E\left\{ u \frac{\big[I + \mu e(t) \mathrm{diag}\big(\vu(t)\big) \big]\vl_a(t)}{\big[\vec{1}^T + \mu e(t) \vu^T(t) \big]\vl_a(t)} \right\} \bigg\Vert^2  \bigg\Vert u \frac{E\left\{\big[I + \mu e(t) \mathrm{diag}\big(\vu(t)\big) \big]\vl_a(t)\right\}}{E\left\{\big[\vec{1}^T + \mu e(t) \vu^T(t) \big]\vl_a(t)\right\}} \bigg\Vert^2 } }
\]
\normalsize
with the same algorithmic parameters as in Fig. \ref{fig:1} and Fig. \ref{fig:2}. We observe that assumptions are fairly accurate for these algorithms in our simulations.

\begin{figure}[t]
\centerline{\epsfxsize=8cm \epsfbox{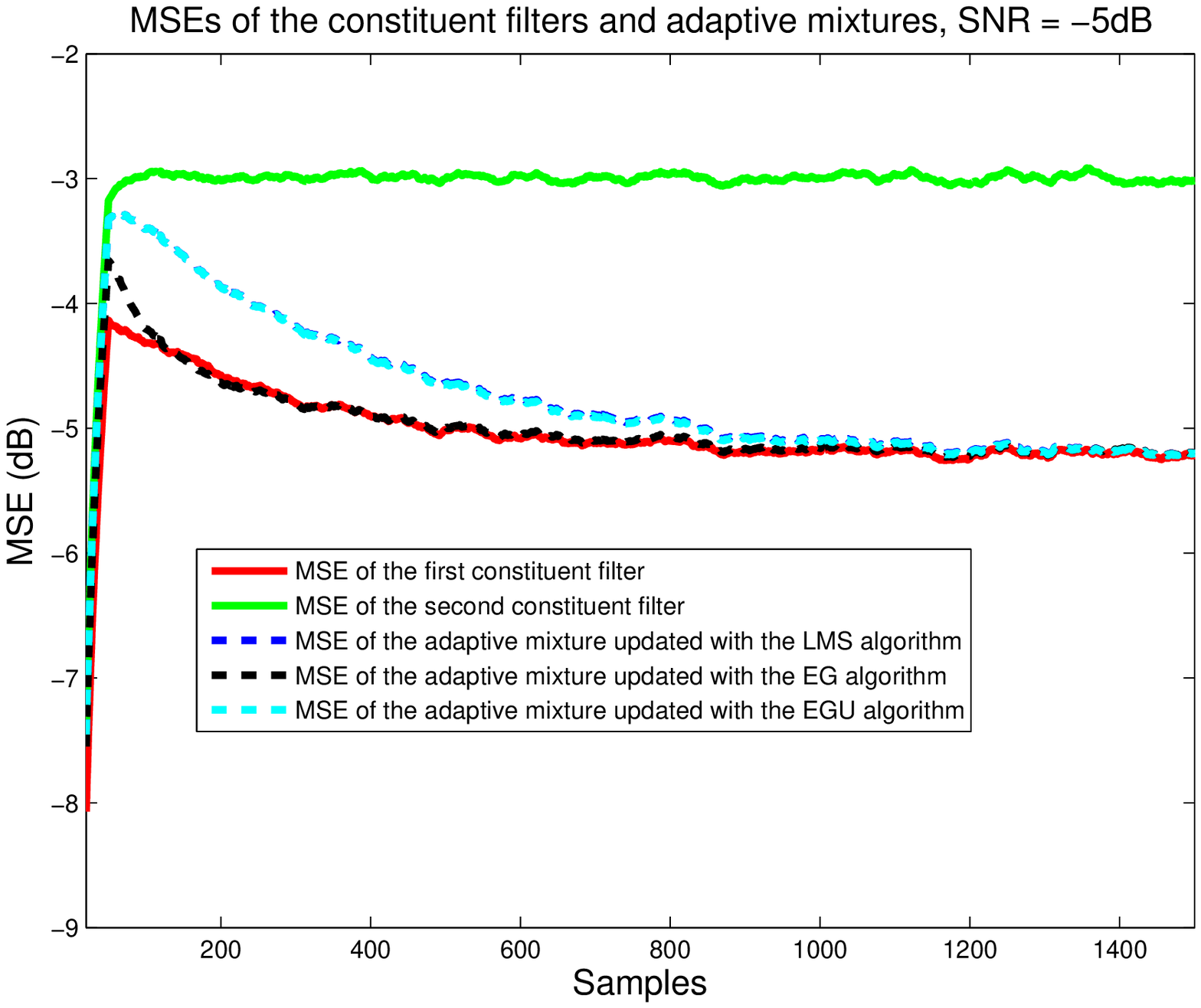}
\hspace{0.2in} \epsfxsize=8cm \epsfbox{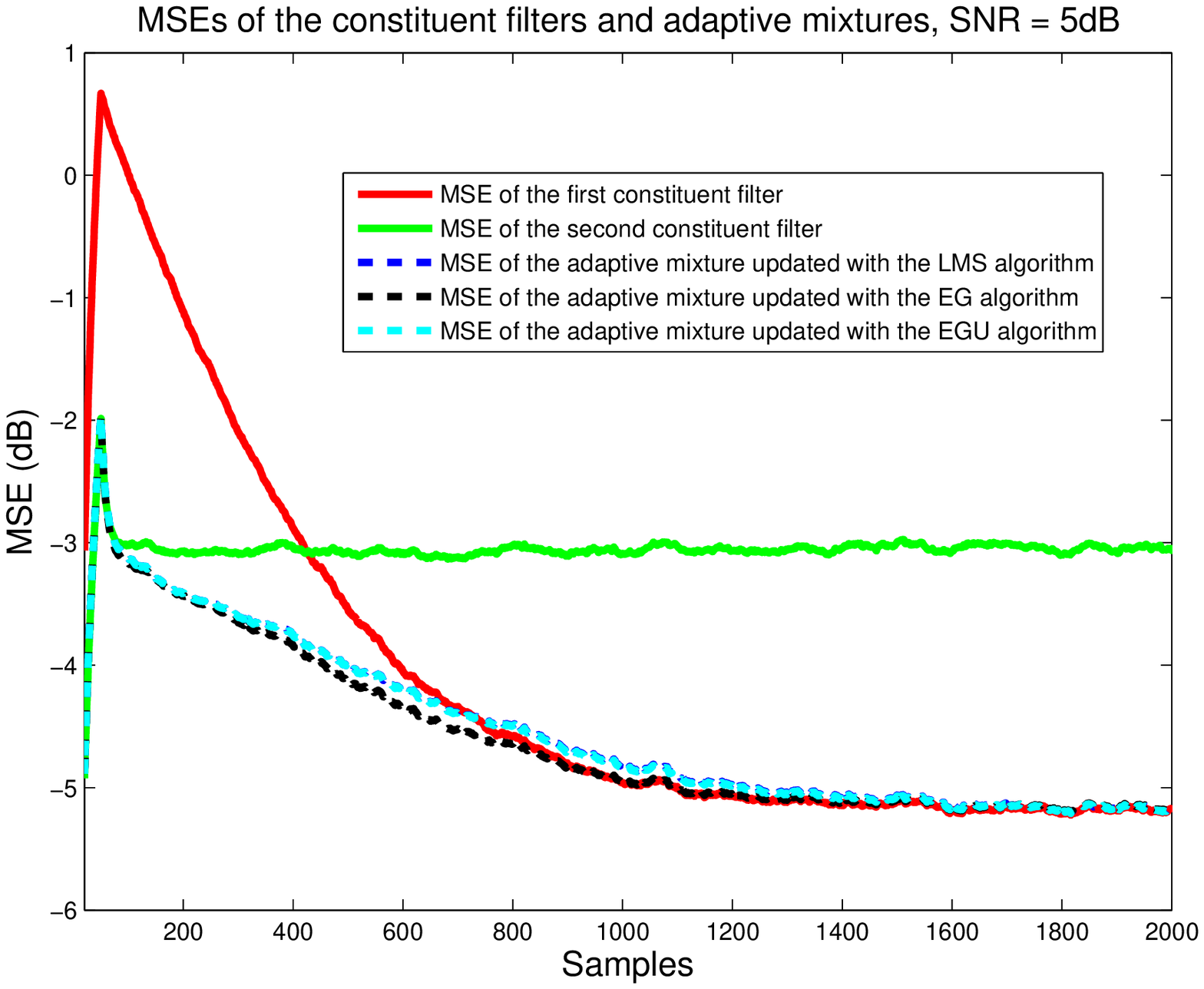} }
\centerline{\hspace{0.96in}(a)\hspace{3.25in}(b)\hfill}
\caption{Algorithmic parameters and constituent filters are selected as in Fig. \ref{fig:1} under SNR = -5dB. For the second stage, the EG algorithm has $\mu_{\mathrm{EG}} = 0.0005$, the EGU algorithm has $\mu_{\mathrm{EGU}} = 0.005$ and the LMS algorithm has $\mu_{\mathrm{LMS}} = 0.005$. For the EG algorithm, u = 500. (a) the MSE curves for the adaptive mixture updated with the EG algorithm, the adaptive mixture updated with the EGU algorithm, the adaptive mixture updated with the LMS algorithm, the first constituent filter and the second constituent filter. Next, SNR = 5dB. For the second stage, the EG algorithm has $\mu_{\mathrm{EG}} = 0.002$, the EGU algorithm has $\mu_{\mathrm{EGU}} = 0.005$ and the LMS algorithm has $\mu_{\mathrm{LMS}} = 0.005$. For the EG algorithm, u = 100. (b) the MSE curves for the adaptive mixture updated with the EG algorithm, the adaptive mixture updated with the EGU algorithm, the adaptive mixture updated with the LMS algorithm, the first constituent filter and the second constituent filter.
\label{fig:4}}
\end{figure}

In the last simulations, we compare performances of the EGU, EG and LMS algorithms updating the affinely mixture weights under different algorithmic parameters. Algorithmic parameters and constituent filters are selected as in Fig. \ref{fig:1} under SNR = -5 and 5. For the second stage, under SNR = -5, learning rates for the EG, EGU and LMS algorithms are selected as $\mu_{\mathrm{EG}} = 0.0005$, $\mu_{\mathrm{EGU}} = 0.005$ and $\mu_{\mathrm{LMS}} = 0.005$ such that the MSEs converge to the same final MSE to provide a fair comparison. We choose u = 500 for the EG algorithm. In Fig. \ref{fig:4}a, we plot the MSE curves for the adaptive mixture updated with the EG algorithm, the adaptive mixture updated with the EGU algorithm, the adaptive mixture updated with the LMS algorithm, first constituent filter with $\mu_1 = 0.002$ and second constituent filter with $\mu_2 \in [0.1,0.11]$ under SNR = -5. Under SNR = 5, learning rates for the EG, EGU and LMS algorithms are selected as $\mu_{\mathrm{EG}} = 0.002$, $\mu_{\mathrm{EGU}} = 0.005$ and $\mu_{\mathrm{LMS}} = 0.005$. We choose u = 100 for the EG algorithm. In Fig. \ref{fig:4}b, we plot same MSE curves as in Fig. \ref{fig:4}a. We observe that the EG algorithm performs better than the EGU and LMS algorithms such that MSE of the adaptive mixture updated with the EG algorithm converges faster than the MSE of adaptive mixtures updated with the EGU and LMS algorithms. We also observe that the EGU and LMS algorithms show similar performances when they are used to train the mixture weights.

\section{Conclusion}

In this paper, we investigate adaptive mixture methods based on Bregman divergences combining outputs of $m$ adaptive filters to model a desired signal. We use the unnormalized relative entropy and relative entropy as distance measures that produce the exponentiated gradient update with unnormalized weights (EGU) and the
exponentiated gradient update with positive and negative weights (EG) to train the mixture weights under the affine constraints or without any constraints. We provide the transient analysis of these methods updated with the EGU and EG algorithms. In our simulations, we compare performances of the EG, EGU and LMS algorithms and observe that the EG algorithm performs better than the EGU and LMS algorithms when the combination vector in steady-state is sparse. We observe that the EGU and LMS algorithms show similar performance when they are used to train the mixture weights. We also observe a close agreement between the simulations and our theoretical results.

\bibliographystyle{elsarticle-num}
\bibliography{msaf_references}

\begin{thebibliography}{10}
\expandafter\ifx\csname url\endcsname\relax
  \def\url#1{\texttt{#1}}\fi
\expandafter\ifx\csname urlprefix\endcsname\relax\def\urlprefix{URL }\fi
\expandafter\ifx\csname href\endcsname\relax
  \def\href#1#2{#2} \def\path#1{#1}\fi

\bibitem{dsp_3}
C.~Boukis, D.~Mandic, A.~G. Constantinides, A class of stochastic gradient
  algorithms with exponentiated error cost functions, Digital Signal Processing
  19 (2009) 201--212.

\bibitem{multip1}
D.~P. Helmbold, R.~E. Schapire, Y.~Singer, M.~K. Warmuth, A comparison of new
  and old algorithms for a mixture estimation problem, Machine Learning 27
  (1997) 97--119.

\bibitem{dsp_1}
J.~C.~M. Bermudez, N.~J. Bershad, J.~Y. Tourneret, Stochastic analysis of an
  error power ratio scheme applied to the affine combination of two lms
  adaptive filters, Signal Processing 91 (2011) 2615--2622.

\bibitem{dsp_2}
J.~Arenas-Garcia, M.~Martinez-Ramon, A.~Navia-Vazquez, A.~R. Figueiras-Vidal,
  Plant identification via adaptive combination of transversal filters, Signal
  Processing 86 (2006) 2430--2438.

\bibitem{kozat2}
S.~S. Kozat, A.~C. Singer, Multi-stage adaptive signal processing algorithms,
  in: Proceedings of SAM Signal Proc. Workshop, 2000, pp. 380--384.

\bibitem{convex3}
J.~Arenas-Garcia, V.~Gomez-Verdejo, M.~Martinez-Ramon, A.~R. Figueiras-Vidal,
  Separate-variable adaptive combination of {L}{M}{S} adaptive filters for
  plant identification, in: Proc. of the 13th IEEE Int. Workshop Neural
  Networks Signal Processing, 2003, pp. 239--248.

\bibitem{convex4}
J.~Arenas-Garcia, M.~Martinez-Ramon, V.~Gomez-Verdejo, A.~R. Figueiras-Vidal,
  Multiple plant identifier via adaptive {L}{M}{S} convex combination, in:
  Proc. of the IEEE Int. Symp. Intel. Signal Processing, 2003, pp. 137--142.

\bibitem{GaGoVi05}
J.~Arenas-Garcia, V.~Gomez-Verdejo, A.~R. Figueiras-Vidal, New algorithms for
  improved adaptive convex combination of lms transversal filters, IEEE
  Transactions on Instrumentation and Measurement 54 (2005) 2239--2249.

\bibitem{EG}
J.~Kivinen, M.~Warmuth, Exponentiated gradient versus gradient descent for
  linear predictors 132 (1997) 1--64.

\bibitem{multip2}
D.~P. Helmbold, R.~E. Schapire, Y.~Singer, M.~K. Warmuth, On-line portfolio
  selection using multiplicative updates, Mathematical Finance 8~(4) (1998)
  325–347.

\bibitem{bershard}
N.~J. Bershad, J.~C.~M. Bermudez, J.~Tourneret, An affine combination of two
  {LMS} adaptive filters: {T}ransient mean-square analysis, IEEE Transactions
  on Signal Processing 56~(5) (2008) 1853--1864.

\bibitem{kozat}
S.~S. Kozat, A.~T. Erdogan, A.~C. Singer, A.~H. Sayed, Steady state {M}{S}{E}
  performance analysis of mixture approaches to adaptive filtering, IEEE
  Transactions on Signal Processing 58 (2010) 4421--4427.

\bibitem{EGvsLMS}
J.~Benesty, Y.~A. Huang, The {LMS},{ PNLMS}, and {E}xponentiated {G}radient
  algorithms, Proc. Eur. Signal Process. Conf. (EUSIPCO) (2004) 721--724.

\bibitem{sayed}
A.~H. Sayed, Fundamentals of Adaptive Filtering, John Wiley and Sons, 2003.

\bibitem{convex}
J.~Arenas-Garcia, A.~R. Figueiras-Vidal, A.~H. Sayed, Mean-square performance
  of a convex combination of two adaptive filters, IEEE Transactions on Signal
  Processing 54 (2006) 1078--1090.

\bibitem{transientconvex}
V.~H. Nascimento, M.~T.~M. Silva, J.~Arenas-Garcia, A transient analysis for
  the convex combination of adaptive filters, IEEE Transactions on Signal
  Processing 58~(8) (2009) 4064--4078.

\bibitem{transientaffine}
S.~S. Kozat, A.~T. Erdogan, A.~C. Singer, A.~H. Sayed, Transient analysis of
  adaptive affine combinations, accepted, 2011.

\bibitem{vovk}
V.~Vovk, A game of prediction with expert advice, Journal of Computer and
  System Sciences 56 (1998) 153--173.

\bibitem{dsp_4}
P.~A. Naylor, J.~Cui, M.~Brookes, Adaptive algorithms for sparse echo
  cancellation, Signal Processing 86 (2006) 1182--1192.

\bibitem{cesab}
N.~Cesa-Bianchi, Y.~Freund, D.~Haussler, D.~P. Helmbold, R.~E. Schapire, M.~K.
  Warmuth, How to use expert advice, Journal of the ACM 44~(3) (1997) 427--485.

\bibitem{dsp_5}
B.~Jelfs, D.~P. Mandic, S.~C. Douglas, An adaptive approach for the
  identification of improper complex signals, Signal Processing 92 (2012)
  335--344.

\end{thebibliography}

\end{document}